\documentclass{article}
\usepackage[letterpaper,top=1in,bottom=1in,left=1in,right=1in,marginparwidth=1.75cm]{geometry}
\usepackage{graphicx}
\usepackage{xcolor}
\usepackage{amssymb,amsmath,amsfonts}
\usepackage{dsfont}
\usepackage{hyperref}
\usepackage{algorithm}
\usepackage{algpseudocode}
\usepackage{hhline}
\usepackage{booktabs}
\usepackage{array}
\usepackage{makecell} 
\usepackage{multirow}
\usepackage{authblk}
\newtheorem{remark}{Remark}

\title{HiPreNets: High-Precision Neural Networks through Progressive Training}

\author{Ethan Mulle\thanks{Department of Applied Mathematics, Baskin School of Engineering, University of California, Santa Cruz; emulle@ucsc.edu.},\ 
Wei Kang\thanks{Professor, Department of Applied Mathematics, Naval Postgraduate School. This work was supported in part by U.S. Naval Research Laboratory - Monterey, CA, the Air Force Office of Scientific Research, USA under Grant No.F4FGA03243G001, and the National Science Foundation (NSF), USA under Grant No.2202668.}, and 
Qi Gong\thanks{Professor, Department of Applied Mathematics, Baskin School of Engineering, University of California, Santa Cruz. This work was supported in part by the National Science Foundation (NSF), USA under Grant No.2134235}}

\begin{document}
\maketitle

\begin{abstract}
Deep neural networks are powerful tools for solving nonlinear problems in science and engineering, but training highly accurate models becomes challenging as problem complexity increases. Non-convex optimization and sensitivity to hyperparameters make consistent performance improvement difficult, and traditional approaches prioritize minimizing mean squared error while overlooking the $L^{\infty}$ norm error that is critical in safety-sensitive applications. To address these challenges, we present HiPreNets, a progressive framework for training high-precision neural networks through sequential residual refinements. Starting from an initial network, each stage trains a refinement network on the normalized residuals of the ensemble so far, systematically reducing both average and worst-case error. A key theme throughout the framework is concentrating training effort on high-error regions of the input domain, which we pursue through complementary techniques including loss function design, adaptive data sampling, localized patching, and boundary-aware training. We validate the framework on benchmark regression problems from the Feynman dataset, where it consistently outperforms standard fully connected networks and reported Kolmogorov-Arnold Networks results, with accuracy approaching machine precision depending on select problems. We further apply the framework to learning the flow map of a 20-dimensional power system ODE, which appears to be the highest dimensional problem studied using this class of multistage methods, achieving substantial reductions in both RMSE and $L^{\infty}$ norm error while enabling a surrogate that predicts system state $238\times$ faster than direct numerical simulation.
\end{abstract}

\section{Introduction}
Fully-connected deep neural networks \cite{Rosenblatt1957,rumelhart1986learning, hornik1989multilayer} have found widespread use in applications such as computer vision \cite{krizhevsky2012imagenet} and natural language processing \cite{attentionvaswani,radford2018improving}, yet they remain underutilized in scientific and engineering domains due to concerns about accuracy and safety \cite{adversarialgoodfellow}. Guaranteeing high accuracy on these problems is difficult because they are often high-dimensional and nonlinear. Much of the existing literature focuses on attaining a ``good enough" average error rather than devoting significant effort to drive it closer to machine precision. Additionally, the final absolute maximum error (which we will refer to throughout this work as the $L^{\infty}$ norm error) of a neural network approximation is generally ignored despite worst-case behavior being critical to analyze for sensitive applications \cite{amodei2016concrete}.

For complex nonlinear, nonconvex problems, it can be hard to know if a neural network has converged to a high quality local minimum, let alone to the global minimum, when optimizing it \cite{choromanska2015loss,dauphin2014identifying}. Once optimization reaches a local minimum or saddle point, escaping to a more optimal solution can be challenging. To do so often requires extensive hyperparameter tuning, which can be time consuming and does not guarantee improvement \cite{bengio2000gradient,zhanggeneralization,snoek2012practical}. Understanding why certain hyperparameter configurations outperform others remains an open challenge. There is therefore generally no structured way of approaching the training and finetuning of a neural network.

One recent line of work explores improving neural network precision and interpretability through changes to model architecture \cite{tan2019efficientnet, kang2022, xie2017aggregated, gong2023105508}. In particular, Kolmogorov–Arnold Networks (KANs) \cite{liu2025kan, KANtime} are a newly proposed architecture inspired by the Kolmogorov–Arnold representation theorem \cite{kolmogorov1957representation}. Instead of relying on compositions of simple, fixed nonlinearities (as in fully connected neural networks), KANs learn the nonlinearity itself through trainable 1D spline functions at each neuron. This gives them greater flexibility and expressivity, particularly for problems with complicated low-dimensional structures. However, KANs introduce new challenges, such as higher memory requirements and the risk of overfitting if not properly regularized.

One classical machine learning approach for increasing the precision is gradient boosting \cite{friedman2001greedy,chen2016xgboost}. Gradient boosting is a machine learning technique used to improve the accuracy of predictions by combining multiple weak models into a stronger model. It does this sequentially, where each new model is trained to correct the errors made by the previous ones. Typically, this has been implemented in practice using decision trees \cite{breicart,quinlan1986induction,quinlan1993c45}, which are individually easy to understand and are low cost to optimize. However, they are prone to overfitting and their outputs are discontinuous and piecewise constant, which can make it difficult to apply them to complex real-world tasks \cite{costa2023recent}.

In recent work, \cite{badirli2020gradient,CHUNG2021113506,michaud2023precision,howard2025} have explored the forms of multi-stage or boosted neural networks. Generally in these methods, fully connected neural networks are used as the ``weak" models to be combined. These models are less interpretable and more costly to optimize than decision trees, but they are much better at capturing nonlinear patterns, can produce robust outputs, and can more easily be regularized to avoid overfitting. In \cite{wang2024multi}, this approach was taken one step further by recognizing neural networks' tendency to learn the low-frequency components of functions first, known as the spectral bias of neural networks \cite{rahaman2019spectral}. As such, the low and high frequency parts of a dataset and the models' residuals can be analyzed and incorporated into the boosting algorithm to further guide the training process. This analysis is insightful and was shown to work for low dimensional problems, but the spectral analysis it relies upon becomes substantially harder to apply as dimensionality grows. In summary, while these earlier works have applied boosting concepts to neural networks, they typically focused on shallow models or spectral insights that are hard to generalize in practice, none of them analyze methods for reducing $L^{\infty}$ norm error, and none of them have investigated high dimensional systems.

To address these issues, we introduce a progressive training framework based on boosting in which the neural network is composed of multiple components trained sequentially. Each component has significantly lower complexity than the full network. This modularity of the framework also offers great flexibility by allowing for each component network to be trained with a different architecture and loss function that may better fit the residuals it encounters. Crucially, a key theme running throughout the framework is the principle of concentrating training effort on high error regions of the input domain. As we will show, this principle can be followed in multiple complementary ways, such as through loss function design, adaptive data sampling, localized patching, and boundary-aware training, with each targeting the same underlying problem of difficult, localized errors that standard training leaves behind. Together, these techniques allow the training to be guided toward reducing either the MSE or the $L^{\infty}$ norm error, depending on the application's requirements. 

We focus exclusively on applying the proposed techniques to regression problems, with accuracy measured by both mean squared error (MSE) and $L^{\infty}$ norm error. While most neural network-based regression methods aim to minimize MSE, often by using it directly as the loss function \cite{rumelhart1986learning}, this approach is suitable only when reducing the average error across the dataset is sufficient. However, it falls short in applications where a single large error could lead to catastrophic failure (e.g., autonomous driving systems or medical diagnostics—see \cite{abrecht2024deep}, \cite{antun2020instabilities}). Our proposed method integrates staged residual refinement with architecture-adaptive modules and explicit $L^{\infty}$ norm error mitigation strategies, enabling highly precise approximations critical in safety-sensitive applications.

In Section \ref{section:algorithm}, we develop the theory behind gradient boosted neural networks in a structured way and analyze the resulting approximation error. We validate the algorithm on benchmark regression problems, investigate the impact of network width on performance, and compare our method against KANs. In Section \ref{section:loss}, we examine how loss function choice affects both RMSE and $L^{\infty}$ norm error, introducing a weighted mean squared error loss in line with the mentioned high error concentration principle. In Sections \ref{section:weightedsampling} and \ref{section:patching}, we introduce weighted sampling and localized patching as further ways to pursue this principle, each concentrating training resources on regions where the current model performs poorly. In Section \ref{section:domainexpand}, we present domain expansion as a boundary-aware technique that addresses a related but distinct source of large errors. In Section \ref{section:highdim}, we apply the framework to a 20-dimensional power system, where targeted data augmentation provides another way to concentrate on large errors, exploiting simulator access to generate new samples in the high error transient regime. Concluding remarks are given in Section \ref{section:conclusion}.

Overall, we validate the proposed method on a family of benchmark regression problems as well as on learning the flow map of a 20-dimensional power system ODE, which appears to be the first application of this style of multistage neural network framework to a problem that high in dimension. (The highest dimension of a problem studied in a previous work appears to be a 6-dimensional synthetic function regression problem in \cite{michaud2023precision}.) Using neural networks to learn flow maps has been an area of great interest in recent years, as summarized in \cite{churchill2023flow}, and doing so in moderately high dimensions is challenging due to nonlinearities and the curse of dimensionality, as analyzed in \cite{brunton2020}. Our results illustrate that the framework scales to such settings, achieving significant reductions in both RMSE and $L^{\infty}$ norm error relative to a base neural network model, while also enabling a neural network surrogate that predicts the system state 5 seconds forward in time $238\times$ faster than a Forward Euler simulator. The code used is available at \url{https://github.com/etmulle/residual-training}.

\section{High-Precision Neural Networks}\label{section:algorithm}
\begin{figure}[h!]
    \centering
    \includegraphics[width=\linewidth]{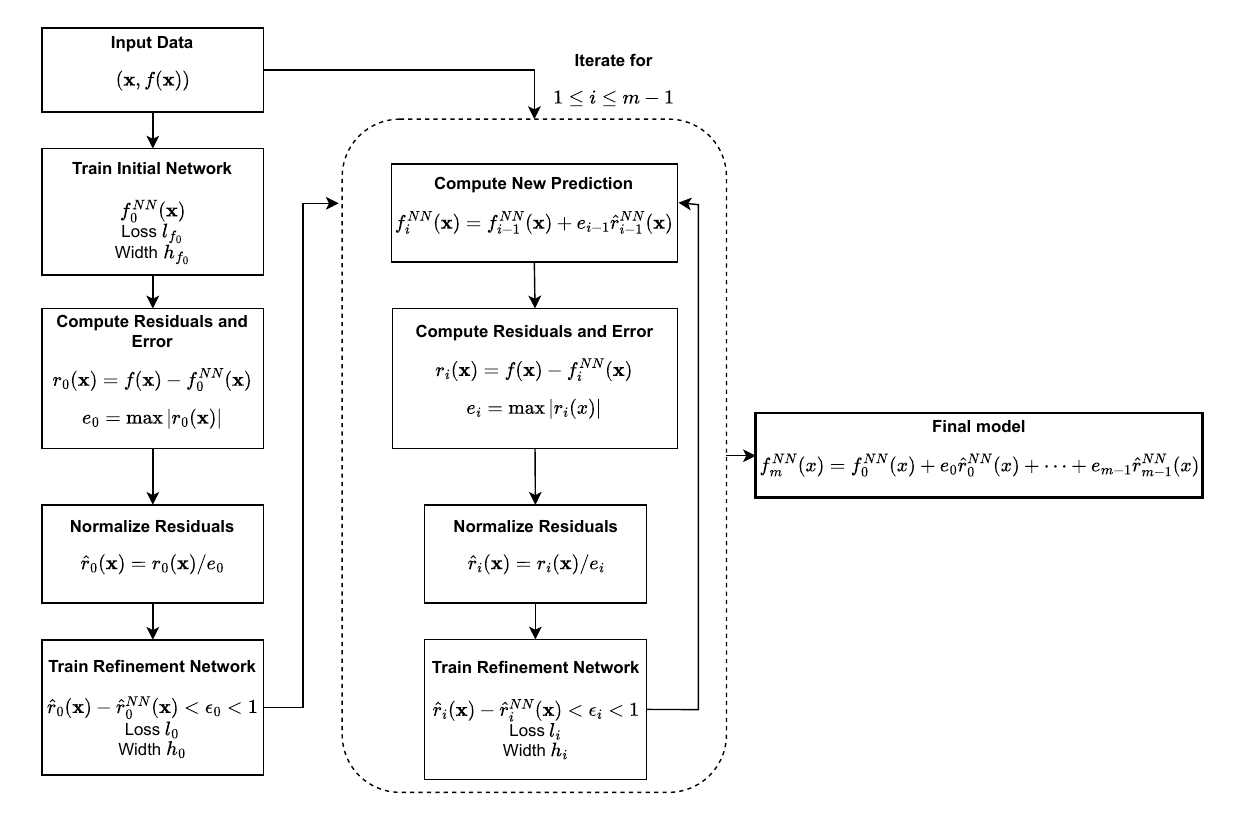}
    \caption{Illustration of the HiPreNet training process. Starting from an initial network $f_0^{NN}$, each subsequent refinement network is trained on the normalized residuals of the ensemble so far, with independent loss function and width. The outputs of all component networks are denormalized and summed to form the final model.}
    \label{fig:model-diagram}
\end{figure}

One of the key advantages of our proposed HiPreNet framework is the breaking down of the haphazard training of a large neural network into the structured and targeted training of several small neural networks. Each of these neural networks will have its own associated loss function $l$ and width $h$. Our framework progressively focuses on errors made by previous learners, enabling each new network to focus on correcting the mistakes of its predecessors. The training process is both simple and intuitive. Once a neural network has been trained and no longer shows significant improvement, we compute the residuals. These residuals are then normalized and used as targets for the next neural network in the sequence.  The predictions of the new network are added to the original model's outputs, yielding a more accurate overall prediction. This process can be repeated multiple times, producing a sequence of refinement networks, each one fine-tuning the ensemble's performance by focusing on remaining errors. A diagram illustrating the training algorithm is given in Figure \ref{fig:model-diagram}.

While the accuracy of an individual refinement network does still depend on chosen hyperparameters, the overall framework contributes significantly to performance gains by systematically addressing model weaknesses. In contrast, the performance of a single large network highly depends on extensive hyperparameter tuning, of which the optimal configuration can vary dramatically across different tasks, making the training process much less predictable. This decentralization of models also contributes to more manageable memory and computational demands, making the training process more accessible and scalable, especially in resource-constrained environments.

\subsection{Algorithmic Formulation of HiPreNet Training}
Let $f(x):D  \to R$ be a continuous function with a compact domain $D \subset R^n$.  In this paper we are interested in approximating the unknown function $f(x)$ based on a given finite dataset of samples $ \{(x_k, f(x_k))\}^N_{k=1}$. We will train a sequence of neural networks $f^{NN}_i(x)$, $i=1,2,\ldots$, on this dataset to approximate the function and make a prediction for any $x \in D$, where the accuracy of the predictions will increase as $i$ increases.

To assess the accuracy of the neural network approximation, we measure the discrepancy between $f(x)$ and its neural network approximation $f^{NN}(x)$ using the $L^p$ norm over domain $D$:
\begin{equation*}
    \|f(x)-f^{NN}(x)\|_{L^p(D)}
\end{equation*}
We first choose $p=2$, where the squared $L^2$ norm is defined as
\begin{equation*}
    \|f(x)-f^{NN}(x)\|^2_{L^2(D)} = \int_D \left(f(x)-f^{NN}(x)\right)^2 dx
\end{equation*}
As we only have a finite amount of samples of $(x,f(x))$, we approximate this integral by the sample average over the dataset and then take the square root to obtain the root mean squared error (RMSE)
\begin{equation*}
    RMSE = \sqrt{\frac{\sum_{i=1}^N \left(f(x_i) - f^{NN}(x_i)\right)^2}{N}}
\end{equation*}
which converges to the true $L_2$ norm as $N\to\infty$.

We additionally consider the $L^{\infty}$ norm defined as
\begin{equation*}
    \|f(x)-f^{NN}(x)\|_{L^{\infty}(D)} = \underset{x \in D}{\max} \left|f(x) - f^{NN}(x)\right|,
\end{equation*}
which is approximated by the maximum prediction error on the given dataset
\begin{equation*}
    \|f(x)-f^{NN}(x)\|_{L^{\infty}(D)} \approx \underset{1\leq i \leq N}{\max} \left|f(x_i) - f^{NN}(x_i)\right|.
\end{equation*}
This metric allows us to evaluate the worst-case behavior of our trained neural network over the domain $D$.

Given a dataset $ \{(x_k, f(x_k))\}^N_{k=1}$, we will design a neural network based model with high approximation accuracy through a progressive training process based on the residuals. Each neural network will be trained using loss function $L$, which, for regression problems, is typically chosen to be MSE loss:
\begin{equation*}
    L_{MSE} = \frac{\sum_{i=1}^N \left(f(x_i) - f^{NN}(x_i)\right)^2}{N}
\end{equation*}

The process initiates with a neural network, $f_0^{NN}(x)$ to approximate $f(x)$. This initial network is usually of small size and low accuracy. 
Denote the initial approximation error $e_0$ as
\begin{equation*}
 e_0  \ = \    \|f(x) - f_0^{NN}(x)\|_{L^{\infty}} .
\end{equation*}

Next we introduce the residuals $r_0(x)$ and the normalized residuals  $\Hat{r}_0(x)$ as
\begin{eqnarray*}
 r_0(x) & = &  f(x) - f_0^{NN}(x), \ \
   \Hat{r}_0(x) \ = \ \frac{r_0(x)}{e_0}.
\end{eqnarray*}
Now, a new neural network,  $\Hat{r}_0^{NN}(x)$,  can be trained to approximate $\Hat{r}_0(x)$ with approximation error $\epsilon_0$ such that 
\begin{equation*}
    \|\Hat{r}_0(x) - \Hat{r}_0^{NN}(x)\|_\infty = \epsilon_0 < 1
\end{equation*}
This generates a new neural network approximation of function $f$ as
\[ f^{NN}_1(x) \ = \ f^{NN}_0(x)+ e_0\Hat{r}_0^{NN}(x),\]
with an approximation error
\begin{eqnarray*}
   e_1 = \|f(x) - f_1^{NN}(x)\|_\infty & = &   \|f_0^{NN}(x) +r_0(x) -  f^{NN}_0(x)- e_0\Hat{r}_0^{NN}(x)\|_\infty\\
   & = &   \|  e_0 \Hat{r}_0(x) - e_0\Hat{r}_0^{NN}(x)\|_\infty\\
   & = &   \epsilon_0 e_0 \ < \ e_0
\end{eqnarray*}
Thus, the new approximation $ f^{NN}_1(x) $ is more accurate than the initial neural network approximation $ f^{NN}_0(x) $.

Repeating this process we can generate a sequence of increasingly accurate neural network approximations of $f(x)$ as 
\begin{eqnarray*}
  f_{i+1}^{NN}(x) & = &  f^{NN}_i(x) + e_i\Hat{r}_i^{NN}(x), 
\end{eqnarray*}
where $e_i =\| f(x) - f_i^{NN}(x)\|_\infty$ is the previous approximation error and $\Hat{r}_i^{NN}$ is a newly trained neural network that approximates, $\Hat{r}_i$, the normalized residual from the previous step, i.e.,
\begin{eqnarray*}
\Hat{r}_i^{NN}\ \approx \ \Hat{ r}_i(x) & = & \frac{ f(x) - f_i^{NN}(x)}{e_i}.
\end{eqnarray*}

\begin{figure}
    \centering
    \includegraphics[width=0.55\linewidth]{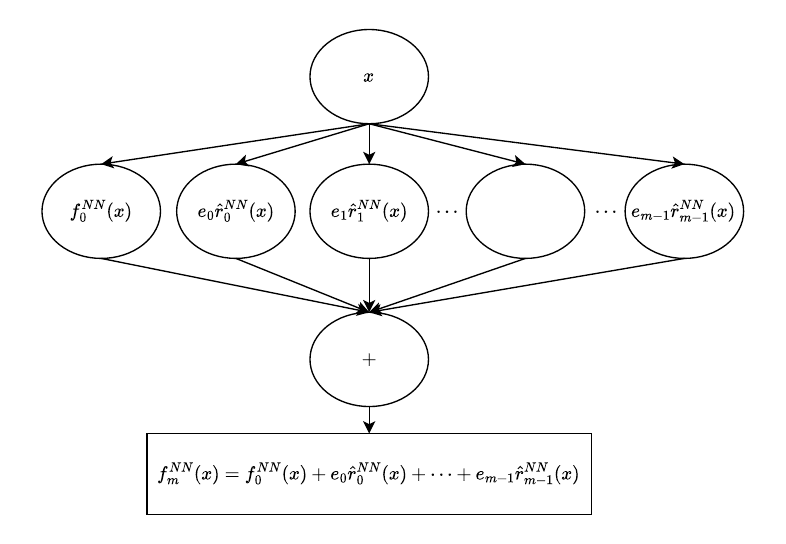}
    \caption{Illustration of making inferences with the trained model. The same input $x$ is the input to all of the trained networks, where their outputs are denormalized and summed together to obtain the final prediction $f_m^{NN}(x)$.}
    \label{fig:evaldiagram}
\end{figure}

Suppose $\Hat{r}_i^{NN}$ is accurate enough so that 
\begin{equation*}
    \|\Hat{r}_i(x) - \Hat{r}_i^{NN}(x)\|_\infty = \epsilon_{i} < 1.
\end{equation*}
Then, we have
\begin{eqnarray*}
   e_{i+1} = \|f(x) - f_{i+1}^{NN}(x)\|_\infty & = &   \|f^{NN}_i(x)+ e_i\Hat{r}_i(x) -  f^{NN}_i(x)- e_i\Hat{r}_i^{NN}(x)\|_\infty\\
   & = &   \|  e_i \Hat{r}_i(x) - e_i\Hat{r}_i^{NN}(x)\|_\infty\\
   & = &   \epsilon_i e_i \ < \ e_i
\end{eqnarray*}
Thus, the accuracy of the new approximation model $ f^{NN}_{i+1}(x) $ keeps increasing with the number of iterations, provided the first neural network approximates the original function with error $e_0 < 1$ and that each refinement neural network approximates the normalized residuals with error $\epsilon_i < 1$. Supposing we have $m$ total refinement networks, this gives a final approximation $f(x)$ as
\begin{equation*}
    f(x) \approx f^{NN}_m(x) = f_0^{NN}(x) + \sum_{i=0}^{m-1}e_i\hat{r}_i^{NN}(x)
\end{equation*}
which is the sum of the outputs of all of the trained model. This is visualized in the diagram given in Figure \ref{fig:evaldiagram}.

As mentioned, the condition $\epsilon_i < 1$ is necessary for each refinement stage to reduce the approximation error. By the Universal Approximation Theorem (UAT) \cite{hornik1989multilayer}, a sufficiently wide network can approximate any continuous function to arbitrary precision, which guarantees that a network achieving $\epsilon_i < 1$ exists in principle for each normalized residual $\hat{r}_i$. However, three practical factors may prevent the training procedure from finding it. First, the residuals become increasingly oscillatory at later stages, and a network of fixed finite width may lack the capacity to approximate them to the required precision, while the UAT requires unbounded width to guarantee arbitrary accuracy. Second, even if a sufficiently expressive network exists, optimization on a non-convex loss landscape may converge to a local minimum with $\epsilon_i \geq 1$. Third, the condition is verified only on the finite training dataset, while the true $L^{\infty}$ norm over the domain may exceed the estimated value $\max|r_i|$, particularly in low-density regions. In all these cases, the reduction in $L^{\infty}$ error may stall or reverse. We observe convergence empirically across all experiments presented in this paper, and treat the framework as an empirically validated algorithm with theoretical motivation rather than a method with provable convergence guarantees. Establishing formal guarantees remains an open direction for future work.

The pseudocode of the algorithm is given in Algorithm \ref{alg:res}.

\begin{algorithm}
    \caption{HiPreNet Training Algorithm}\label{alg:res}
    \begin{algorithmic}
        \Require Dataset $(\textbf{x}, \textbf{f}) = \{(x_k, f(x_k))\}^N_{k=1}$, trained neural network $f_{0}^{NN}$, tolerance $tol$
        \State $error \gets L\left(\textbf{f}, f_{0}^{NN}(\textbf{x})\right)$
        \State $r_0(\textbf{x}) \gets \textbf{f} - f_{0}^{NN}(\textbf{x})$
        \State $e_0 \gets \max{|r_0(\textbf{x})|}$ \Comment{estimate of true $e_0$}
        \State $\Hat{r}_0(\textbf{x}) \gets \frac{r_0}{e_0}$
        \State  $\Hat{r}^{NN}_0(\textbf{x}) \gets train(\textbf{x},\Hat{r}_0(\textbf{x}))$
        \State $i \gets 1$
        \While{$error > tol$}
            \State  $f^{NN}_{i}(\textbf{x}) \gets f_{i-1}^{NN} + e_{i-1} \Hat{r}_{i-1}^{NN}(\textbf{x})$
            \State $r_{i}(\textbf{x}) \gets \textbf{f} - f^{NN}_{i}(\textbf{x})$
            \State $e_{i} \gets \max{|r_i(\textbf{x})|}$ \Comment{estimate of true $e_{i}$}
            \State $\Hat{r}_{i}(\textbf{x}) \gets \frac{r_{i}(\textbf{x})}{e_{i}}$
            \State $\Hat{r}^{NN}_{i}(\textbf{x}) \gets train(\textbf{x}, \Hat{r}_{i}(\textbf{x}))$
            \State $error \gets L\left(\textbf{f}, f_{i}^{NN}(\textbf{x})\right)$
            \State $i \gets i + 1$
        \EndWhile
    \end{algorithmic}
\end{algorithm}

\begin{remark}
The total size of the model depends directly on the size of the neural networks used in each training iteration. Additionally, the architecture of the network used in each iteration does not have to remain fixed; the loss function, activation function, and number of layers and neurons used in the first iteration can be completely different from those used for the last iteration. This allows for increased adaptability to what is observed in the residuals, where different architectures may be better suited for different residual patterns.
\end{remark}

\begin{remark}
It is important to note that the true approximation error $e_i$ cannot be obtained in practice, as it is defined using the $L^{\infty}$ norm and we can only train the neural networks on a limited dataset. As such, in our algorithm, we estimate $e_0$ from residuals $r_0$ as
\begin{equation*}
    e_0 \approx \max |r_0|
\end{equation*}
and each $e_i$ from residuals $r_i$ as
\begin{equation*}
    e_i \approx \max|r_i(\textbf{x})|
\end{equation*}
\end{remark}

\subsection{Preliminary Experiments on a Benchmark Function}
Throughout the following sections, we test our algorithm on five different functions from the Feynman dataset first introduced in \cite{aifeynman2020}. Each function is numbered and corresponds to a different dimensionless multi-variable function. These functions correspond to well known physical laws and relations. The functions are written out in Table \ref{tab:equations} with their number in the dataset and both their original and dimensionless forms. For this paper, we attempt to learn the dimensionless forms of the functions. The dataset contains 1,000,000 samples for each function, which were all standardized and used for training the networks. To test the networks, we randomly generate an additional 1,000,000 validation samples for each function with each variable in the domains given in \cite{aifeynman2020}.
\begin{table}[h]
    \centering
    \renewcommand{\arraystretch}{1.8}
    \begin{tabular}{c>{\centering\arraybackslash}m{5cm}>{\centering\arraybackslash}m{5cm}c}
        \hline
        \textbf{Number} & \textbf{Function} & \textbf{Dimensionless Function} & \textbf{Dimensionality}\\
        \hline
        I.6.2 & $f(\theta,\sigma)=\exp{(-\frac{\theta^2}{2\sigma^2})}/\sqrt{2\pi \sigma^2}$ & $f(\theta,\sigma)=\exp{(-\frac{\theta^2}{2\sigma^2})}/\sqrt{2\pi \sigma^2}$ & 2\\
        I.9.18 & $f(G,\textbf{m},\textbf{x},\textbf{y},\textbf{z})= \frac{G m_1 m_2}{(x_2 - x_1)^2 + (y_2 - y_1)^2 + (z_2 - z_1)^2}$ & $f(a,b,c,d,e,f)=\frac{a}{(b-1)^2 + (c-d)^2 + (e-f)^2}$ & 6\\
        I.13.12 & $f(G,m_1,m_2,r_1,r_2)=G m_1 m_2 (\frac{1}{r_2} - \frac{1}{r_1})$ & $f(a,b)=a(\frac{1}{b} - 1)$ & 2\\
        I.26.2 & $f(n,\theta_2)=\arcsin{(n\sin \theta_2)}$ & $f(n,\theta_2)=\arcsin{(n\sin \theta_2)}$ & 2\\
        I.29.16 & $f(x_1,x_2,\theta_1,\theta_2)=\sqrt{x_1^2 + x_2^2 - 2 x_1 x_2 \cos{(\theta_1 - \theta_2)}}$ & $f(a,\theta_1,\theta_2)=\sqrt{1 + a^2 - 2a\cos{(\theta_1 - \theta_2)}}$ & 3\\
        \hline
    \end{tabular}
    \caption{Functions selected from the Feynman dataset, shown with their original physical forms, dimensionless forms, and input dimensionality used in experiments.}
    \label{tab:equations}
\end{table}

To begin with, we first compare our algorithm's performance on Function I.13.12 with the performance of a single standard fully connected network with the same number of parameters, but made wider (going from 5 to 11 neurons in each hidden layer) or deeper (going from 5 to 19 hidden layers of 5 neurons each). For our algorithm, an initial predictor neural network as well as three refinement networks were initialized and trained. All networks contain five hidden layers with five neurons each and use tanh as their activation function.  

Each refinement network is optimized with SciPy's \cite{2020SciPy-NMeth} Broyden–Fletcher–Goldfarb–Shanno (BFGS) algorithm \cite{nocedal2006numerical} for a maximum of 25000 iterations, using a gradient tolerance of 1e-12. The wide and deep fully connected networks are optimized with BFGS for a maximum of 100000 iterations. We use BFGS for training each network due to its strong local convergence properties and robustness in high precision settings. As a quasi-Newton method, BFGS leverages curvature information to guide updates more effectively than first-order methods, enabling more reliable convergence to low error solutions. While computationally more intensive due to scaling as $O(n^2)$ in memory, its performance is well-suited to the relatively small networks used in our framework, making it an ideal choice for achieving the extremely low errors targeted for these low dimensional functions.

We compare the lowest RMSE over three independent trials, solely changing the random seed. As shown in Table \ref{tab:I1312_metrics}, the HiPreNet approach achieves a lower RMSE than either of the traditional fully connected neural network approaches.
\begin{table}[h!]
    \centering
     \renewcommand{\arraystretch}{1.2}
    \begin{tabular}{lccc}
        \hline
        \textbf{Method} &  \textbf{Validation RMSE} & \textbf{Parameters per Training} & \textbf{Size of Model} \\
        \hline
        \textbf{Wider FCNN}   & 6.851e-06 & 573 & 573 \\
        \textbf{Deeper FCNN}   & 1.485e-04 & 561 & 561 \\
        \textbf{HiPreNet}  & 2.611e-06 & 141 & 564 \\
        \hline
    \end{tabular}
    \caption{Comparison of validation RMSE and parameter counts for wider, deeper, and HiPreNet configurations on Function I.13.12.}
    \label{tab:I1312_metrics}
\end{table}

\subsection{Comparison with Kolmogorov–Arnold Networks (KANs)}
It is admittedly not difficult to beat an untuned fully connected network in performance, so we also compare our results to the results obtained by the Kolmogorov-Arnold networks (KANs) and fully connected networks presented in \cite{liu2025kan}, which are evaluated on the same Feynman dataset. As previously mentioned, instead of having fixed activation functions like standard fully connected networks, KANs learn the nonlinearities themselves through trainable 1D spline functions at each neuron.

For each Feynman function dataset, an initial predictor neural network as well as three refinement networks were initialized and trained. All networks contain five hidden layers with five neurons each and use tanh as their activation function. Each network was trained using BFGS as the optimizer with 25000 maximum iterations. We run each simulation for three different random seeds and report the results for the seed that provided the lowest validation error.

The initial RMSE and final RMSE after the training the HiPreNets are compared to each other for the different validation datasets in Table \ref{tab:metrics_table}. We also compare the number of parameters needed for each model. As previously mentioned, in training a HiPreNet, multiple neural networks are trained sequentially and are then combined to create a final model for inference. In Table \ref{tab:metrics_table}, the ``Parameters per Training" column indicates the number of parameters in each individual neural network, while the ``Total Model Size" column indicates the total number of parameters in the final model. It is not precisely clear the exact number of parameters used in each KAN from \cite{liu2025kan}, so we report an estimate based on the authors' presented figures in the ``KAN Model Size" column. Following from this, we note that this comparison is approximate, as the KAN results are taken directly from \cite{liu2025kan} rather than a re-implementation, and differences in aspects such as the training budget and optimizer settings may exist. Nonetheless, the comparison is still informative as an order-of-magnitude benchmark against a recent advanced architecture evaluated on the same dataset.

\begin{figure}[h!]
    \centering
    \includegraphics[width=16cm]{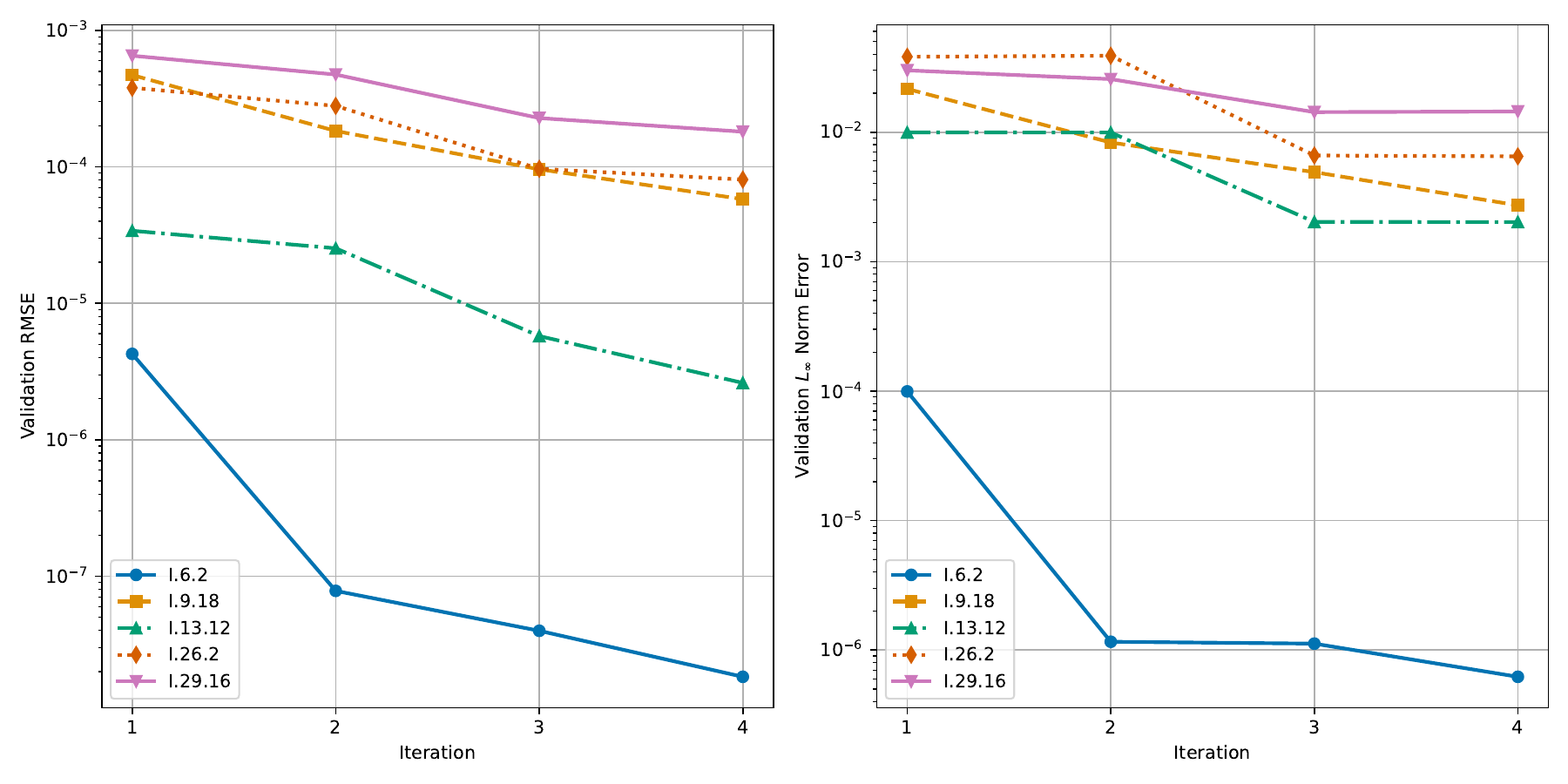}
    \caption{RMSE (a) and $L^{\infty}$ norm error (b) at each stage of HiPreNet training for the five Feynman benchmark functions. Iteration 1 corresponds to the initial network, and each subsequent iteration adds one residual refinement network. While RMSE decreases consistently across all functions, $L^{\infty}$ norm error shows less uniform improvement.}
    \label{fig:initialsimulations}
\end{figure}

The change in RMSE as more refinement networks are trained is visualized in Figure \ref{fig:initialsimulations}(a). In all cases, each refinement network successively reduces the error. By the addition of the final refinement network, the final RMSE obtained is lower than the best reported RMSE in \cite{liu2025kan}. For Function I.6.2 in particular, by the addition of the final refinement network, the MSE (1.832e-08 RMSE = 3.356e-16 MSE) has reached near machine epsilon (2.220e-16) for 64-bit double precision. 

\begin{table}[h!]
    \centering
     \renewcommand{\arraystretch}{1.2}
    \begin{tabular}{ccccccc}
        \hline
        \textbf{Function} & \makecell{\textbf{Initial} \\ \textbf{RMSE}} & \makecell{\textbf{Final} \\ \textbf{RMSE}}
         & \makecell{\textbf{KAN} \\ \textbf{RMSE}}    & \makecell{\textbf{Parameters}  \\ \textbf{per Iteration}} 
         & \makecell{\textbf{Total} \\ \textbf{Model Size}} & \makecell{\textbf{KAN} \\ \textbf{Model Size}} \\
        \hline
        \textbf{I.6.2}   & 4.262e-06 & 1.832e-08 & 2.86e-05 & 141 & 564 & $\mathcal{O}(10^2)$ \\
        \textbf{I.9.18}  & 4.712e-04 & 5.801e-05 & 1.48e-03 & 161 & 644 & $\mathcal{O}(10^2)$ \\
        \textbf{I.13.12} & 3.403e-05 & 2.611e-06 & 1.42e-03 & 141 & 564 & $\mathcal{O}(10^2)$ \\
        \textbf{I.26.2}  & 3.821e-04 & 8.076e-05 & 7.90e-04 & 141 & 564 & $\mathcal{O}(10^2)$ \\
        \textbf{I.29.16} & 6.528e-04 & 1.811e-04 & 3.20e-03 & 146 & 584 & $\mathcal{O}(10^2)$ \\
        \hline
    \end{tabular}
    \caption{Validation RMSE, per-iteration parameter count, and total model size for HiPreNets versus KANs across all five Feynman functions. KAN results are reproduced from \cite{liu2025kan}. HiPreNets achieve lower RMSE than the reported KAN results in all cases.}
    \label{tab:metrics_table}
\end{table}

\subsection{Empirical Insights from Early Experiments}
The previous results have demonstrated that the HiPreNet framework provides a structured way of reducing the RMSE on a set of problems as compared to both traditional approaches of using fully connected neural networks and more recent KANs. To further understand the advantages and disadvantages of the HiPreNet framework, we first report the $L^{\infty}$ norm error on the five chosen Feynman functions in Table \ref{tab:linf_norms_table}. We also show the change in $L^{\infty}$ norm error as more refinement networks are trained in Figure \ref{fig:initialsimulations}(b). As previously mentioned, the $L^{\infty}$ norm error is a valuable metric for quantifying the worst-case behavior of a model, as opposed to RMSE which quantifies the average behavior of a model. Analyzing this behavior is critical for applications such as healthcare, where a catastrophic single error can have enormous real-world consequences. Although there is a reduction in $L^{\infty}$ norm error by the time the final refinement network is added, it is less noticeable than the reduction in RMSE.
\begin{table}[h!]
    \centering
     \renewcommand{\arraystretch}{1.2}
    \begin{tabular}{ccc}
        \hline
        \textbf{Function} & \textbf{Initial $L^{\infty}$ Norm} & \textbf{Final $L^{\infty}$ Norm} \\
        \hline
        \textbf{I.6.2}   & 9.948e-05 & 6.221e-07 \\
        \textbf{I.9.18}  & 2.157e-02 & 2.722e-03 \\
        \textbf{I.13.12} & 9.941e-03 & 2.021e-03 \\
        \textbf{I.26.2}  & 3.822e-02 & 6.527e-03 \\
        \textbf{I.29.16} & 3.000e-02 & 1.443e-02 \\
        \hline
    \end{tabular}
    \caption{Initial and final $L^{\infty}$ norm errors for each function, showing the reduction achieved by the HiPreNet training procedure.}
    \label{tab:linf_norms_table}
\end{table}

\begin{figure}[h!]
    \centering
    \includegraphics[width=1\linewidth]{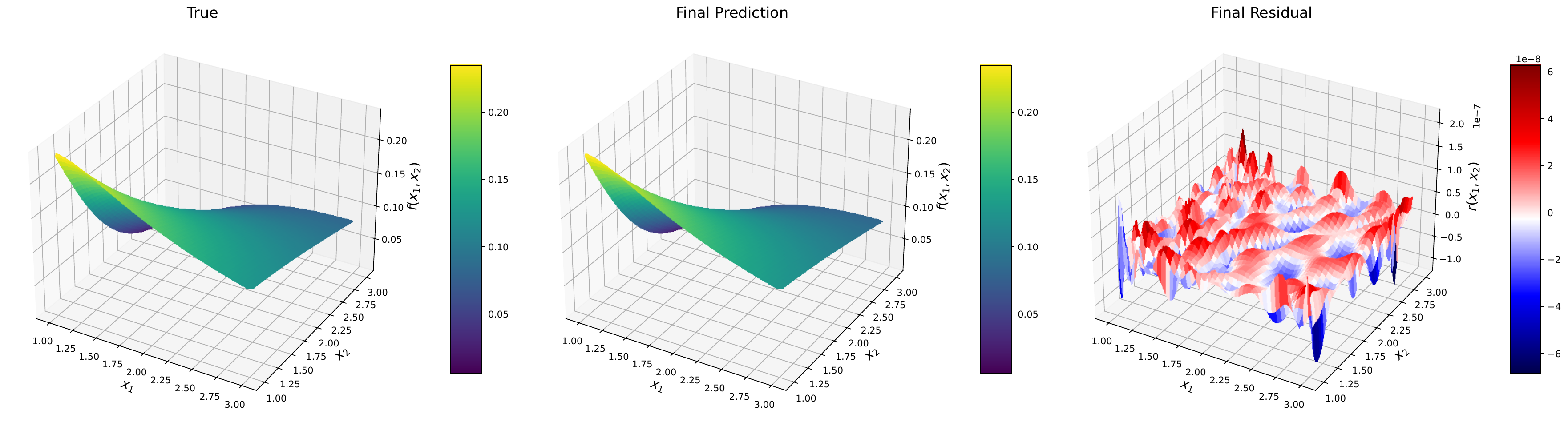}
    \caption{Visual comparison of the true function for Function I.6.2, the final model approximation, and the remaining residuals after training. The residuals exhibit peaks concentrated along the domain boundary.}
    \label{fig:surfaceinitialfinal}
\end{figure}

\begin{figure}
    \centering
    \includegraphics[width=0.9\linewidth]{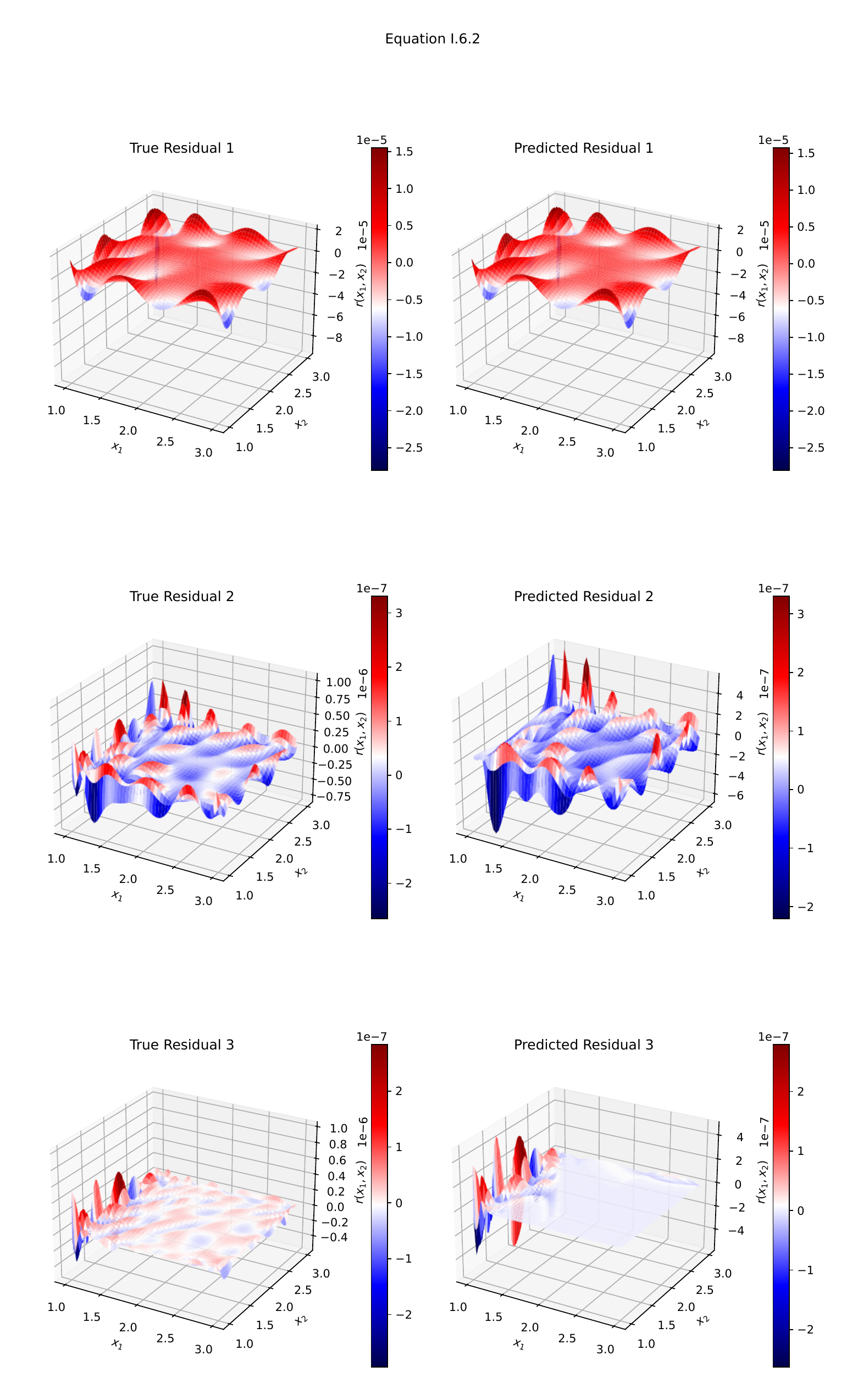}
    \caption{Surface plots of the true and predicted residuals at each stage of HiPreNet training for Function I.6.2, with rows corresponding to successive refinement stages. Note that axis and colorbar scales differ across plots. The overall residual magnitude decreases with each stage, though later residuals exhibit increasingly oscillatory structure that becomes harder to approximate.}
    \label{fig:initialresidualsurface}
\end{figure}

We also further illustrate the results for Function I.6.2, though the following observation holds true for all of the functions in Table \ref{tab:equations}. In Figure \ref{fig:surfaceinitialfinal} we show plots of the true function, the final approximation after training a HiPreNet, and the final residuals. In Figure \ref{fig:initialresidualsurface}, we show surface and contour plots of the residuals at each stage of the training process. Each successive network generally succeeds in approximating the residuals of its predecessor on average, but has difficulty in capturing the more local, oscillatory and high frequency behavior that emerges with the later residuals, often along the boundary of the dataset. This indicates that the later networks may not be powerful enough to precisely capture this more complex behavior. When this is combined with the choice of MSE loss which focuses on reduction of the average error, there is little incentivizing a decrease in the $L^{\infty}$ norm error. 

Therefore, while the HiPreNet algorithm has already shown to be a viable method for reducing RMSE in a structured way, there are three key takeaways of this initial approach that we take note of and will investigate in the coming sections to further improve the algorithm:
\begin{enumerate}
    \item The residual functions become more oscillatory as more networks are added. Early residuals show coarse, low frequency errors, while later residuals often display more localized and higher frequency behavior and are therefore more difficult to learn. Knowing this, how many parameters are needed in each successive network?
    \item Using mean squared error as the loss function for the refinement networks does reduce the average error, but is less ideal for reducing the maximum ($L^{\infty}$ norm) error. Is there a better loss function that can be used to prioritize reducing the $L^{\infty}$ norm error?
    \item Before training each network, we can analyze the residuals to observe any regions that are difficult to learn. Can we modify the HiPreNet algorithm to take advantage of this information? 
\end{enumerate}

\section{Network Width Adaptation in Residual Stages}\label{section:width}
From classical neural network theory it is known that as the width of a fully connected neural network increases, the complexity of the behavior that it can capture grows. As we have observed that the complexity of the residuals grow as more iterations of the HiPreNet algorithm happen, this indicates that the width of the refinement networks should be consistently increased to match this. 

An empirical observation throughout our experiments is that the residual functions at each stage tend to become increasingly oscillatory, often exhibiting a greater number of local extrema. This suggests that the complexity of the residuals grows with each iteration, possibly in a linear fashion. One potential heuristic for determining the width of successive refinement networks is to match the network capacity to the number of extrema or local variations in the residuals. While we adopt a linearly increasing width schedule in this work, a more principled approach remains an open area for future investigation.

To confirm that this idea improves the HiPreNet framework, we fix the initial neural network and three refinement neural networks to all have a depth of five hidden layers. We then vary the width of the layers for each network to be [5-5-5-5],  [20-20-20-20], and [5-10-15-20]. For example, for the [5-10-15-20] case, the initial network will have 5 neurons in each layer, the first refinement network will have 10 neurons in each layer, the second refinement network will have 15, and the third refinement network will have 20. This allows for each additional network to have more capacity to fit the increasingly complex residual structures.

\begin{figure}[h!]
    \centering
    \includegraphics[width=16.5cm]{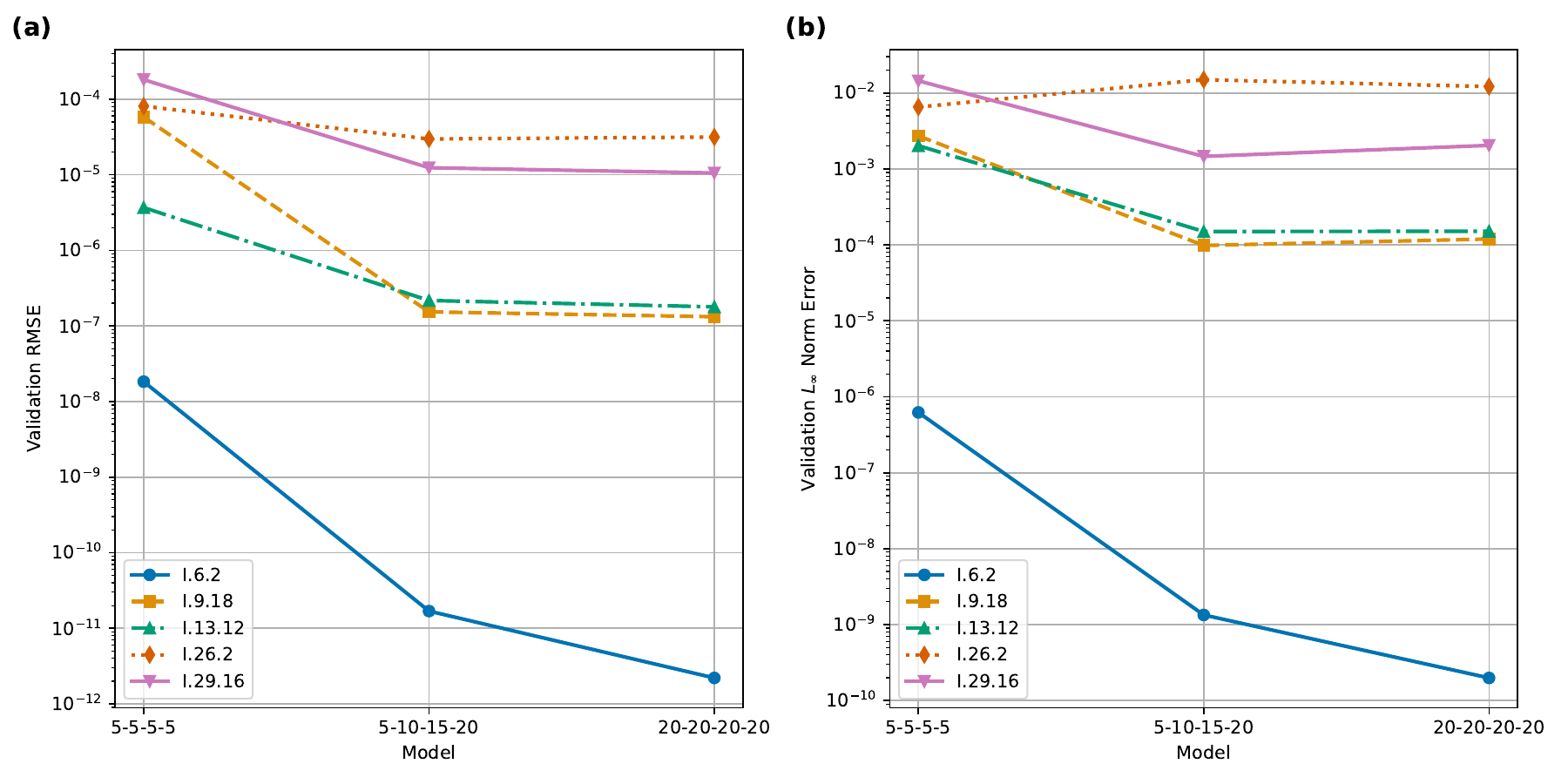}
    \caption{Validation RMSE (a) and $L^{\infty}$ norm error (b) for all five Feynman functions under three HiPreNet width configurations: uniform [5-5-5-5], linearly increasing [5-10-15-20], and uniform [20-20-20-20]. Linearly increasing widths generally achieve comparable or better accuracy than the [20-20-20-20] configuration at roughly half the total parameter count.}
    \label{fig:numneurons}
\end{figure}

\begin{figure}
    \centering
    \includegraphics[width=0.7\linewidth]{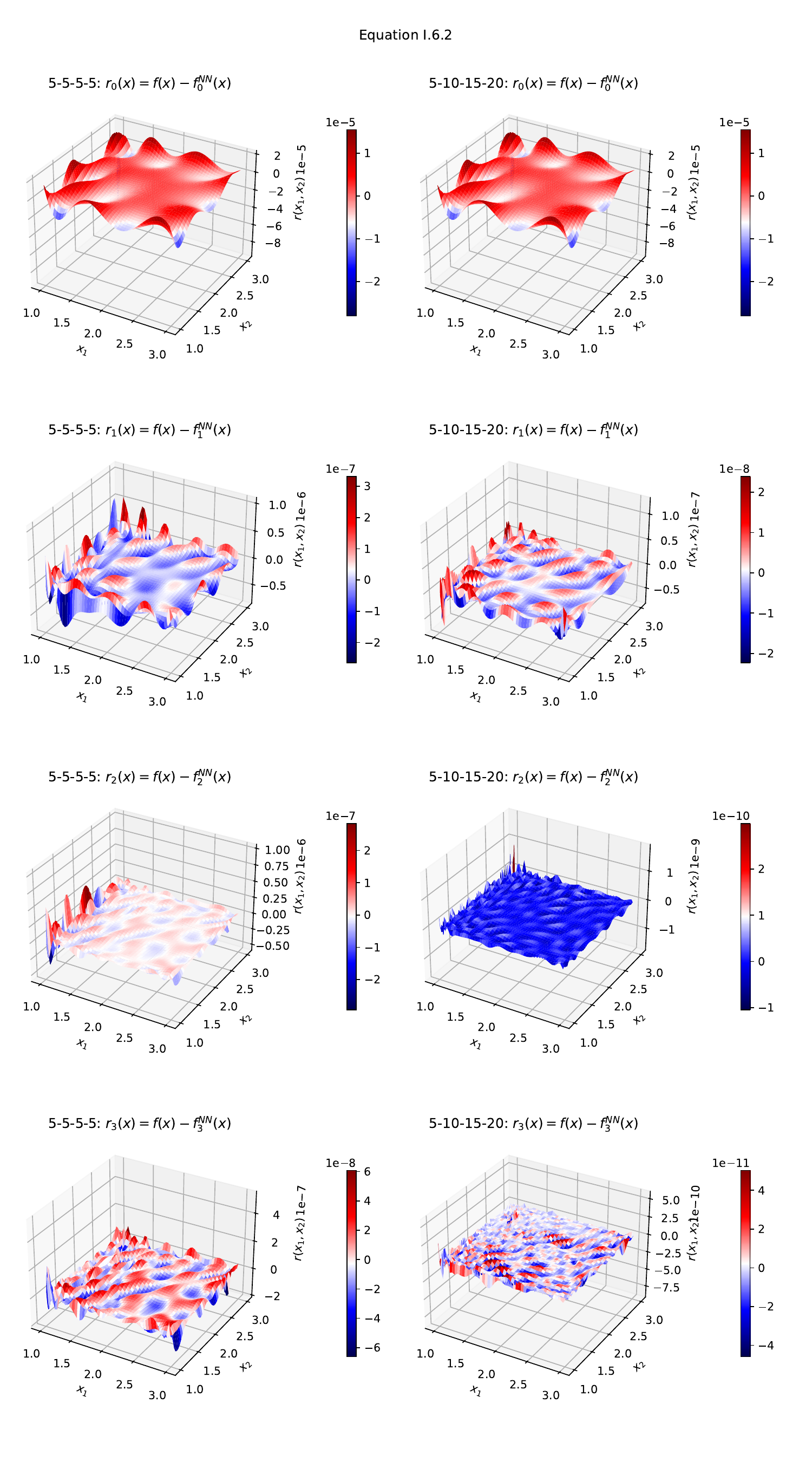}
    \caption{Side-by-side comparison of residuals at each training stage for Function I.6.2, with the [5-5-5-5] configuration (left column) and the [5-10-15-20] configuration (right column). Note that axis and colorbar scales differ across plots. The linearly increasing width configuration achieves substantially smaller residuals by the later stages, reflecting the increased capacity available as residual complexity grows.}
    \label{fig:numneurons_surfaces}
\end{figure}

The results are summarized in Table \ref{tab:neurons_metrics_table} and plotted in Figure \ref{fig:numneurons}. We first observe that for the HiPreNet approach, there is a significant reduction in RMSE and generally a smaller reduction in $L^{\infty}$ norm error (with the exception of Function I.26.2) when moving from a [5-5-5-5] setup to [5-10-15-20], corresponding to roughly a sixfold increase in the total number of parameters. A comparison of the residuals at each iteration when using [5-5-5-5] versus [5-10-15-20] is shown in Figure \ref{fig:numneurons_surfaces}.
\begin{table}[h!]
    \centering
       \renewcommand{\arraystretch}{1.5}
    \begin{tabular}{cccccc}
        \hline
         \textbf{Function} & \makecell{\textbf{Network} \\ \textbf{Structure}} 
          & \makecell{\textbf{Final} \\ \textbf{RMSE}} 
         & \makecell{\textbf{Final $L^{\infty}$} \\ \textbf{Norm Error}}
         & \makecell{\textbf{Parameters} \\ \textbf{per Iteration}}    
         & \makecell{\textbf{Model} \\ \textbf{Size}} \\
        \hline
        \multirow{3}{*}{\textbf{I.6.2}} & 5-5-5-5   & 1.832e-08 & 6.221e-07 & 141 & 564 \\
         & 20-20-20-20   & 2.204e-12 & 1.987e-10 & 1761 & 7044 \\
        & 5-10-15-20   & 1.688e-11 & 1.341e-09 & 141, 481, 1021, 1761 & 3404 \\
        \hline
         \multirow{3}{*}{\textbf{I.9.18}} &  5-5-5-5  & 5.801e-05 & 2.722e-03 & 161 & 644 \\
         & 20-20-20-20  & 1.324e-07 & 1.194e-04 & 1841 & 7364\\
         &  5-10-15-20   & 1.535e-07 & 9.822e-05 & 161, 521, 1081, 1841 & 3604\\
        \hline
         \multirow{3}{*}{\textbf{I.13.12}}& 5-5-5-5 & 2.611e-06 & 2.021e-03 & 141 & 564  \\
         & 20-20-20-20 & 1.789e-07 & 1.510e-04 & 1761 & 7044 \\
         & 5-10-15-20 & 2.171e-07 & 1.501e-04 & 141, 481, 1021, 1761 & 3404  \\
        \hline
        \multirow{3}{*}{ \textbf{I.26.2}} &  5-5-5-5  & 8.076e-05 & 6.527e-03 & 141 & 564 \\
         & 20-20-20-20  & 3.156e-05 & 1.214e-02 & 1761 & 7044\\
         & 5-10-15-20  & 2.977e-05 & 1.495e-02 & 141, 481, 1021, 1761 & 3404\\
        \hline
        \multirow{3}{*}{ \textbf{I.29.16}} & 5-5-5-5 & 1.811e-04 & 1.443e-02 & 146 & 584\\
         &  20-20-20-20 & 1.056e-05 & 2.040e-03 & 1781 & 7124\\
         &  5-10-15-20 & 1.239e-05 & 1.461e-03 & 146, 491, 1036, 1781 & 3454\\
        \hline
    \end{tabular}
    \caption{Validation RMSE and $L^{\infty}$ norm error for three HiPreNet width schedules ([5-5-5-5], [5-10-15-20], [20-20-20-20]) across all five functions, with per-iteration and total parameter counts (model size).}
    \label{tab:neurons_metrics_table}
\end{table}


One might also expect that simply making each network bigger will lead to a reduction in loss. However, increasing the setup further from [5-10-15-20] to [20-20-20-20], which approximately doubles the parameter count, yields minimal improvement in both RMSE and $L^{\infty}$ norm error. This suggests that simply setting all widths to the same value is less efficient than gradually increasing them. Moreover, in the [20-20-20-20] case, the later networks may struggle to capture the complex residual behavior from their preceding network, indicating that a schedule like [20-25-30-35] might be needed. However, this quickly becomes computationally infeasible, requiring increasingly more memory and compute time especially if using BFGS as the optimizer. Overall, gradually widening the refinement networks proves to be a more effective strategy for balancing computational cost with model expressiveness, and careful design of the number of parameters in each neural network provides an important avenue for enhancing final model performance.

\section{$L^{\infty}$ Norm Error Reduction}
\subsection{Loss Function Design for $L^{\infty}$ Norm Error Reduction} \label{section:loss}
As previously discussed, when using mean-squared error as the loss function for the refinement networks, the $L^{\infty}$ error may remain constant or even increase despite the mean-squared error consistently decreasing. This is often due to sharp peaks in the residuals, which are typically located near the boundaries of the input domain. Using MSE as the loss function may cause the refinement networks to ignore these extreme, but very localized deviations in favor of focusing on areas which better reduce the mean error.

Direct optimization of the $L^{\infty}$ norm is difficult, as it is non-differentiable and highly sensitive to single-point fluctuations, which complicates the training dynamics of neural networks. Moreover, relatively little prior work has focused on developing loss functions that specifically aim to minimize the $L^{\infty}$ error. One way we attempt to focus the refinement networks on reducing the $L^{\infty}$ norm error is by using a weighted mean-squared error (WMSE) loss function. If at iteration $i$ we are training refinement network $\hat{r}^{NN}_i$ to approximate the residual $\hat{r}_i$, the WMSE loss function is defined as
\begin{equation*}
    WMSE = \frac{\sum_{k=1}^N w_k\left(\hat{r}_i(x_k) - \hat{r}_i^{NN}(x_k)\right)^2}{N}
\end{equation*}
where
\begin{equation*}
    w_k = \left(1+|\hat{r}_i(x_k)|\right)
\end{equation*}
Note that for this approach, the loss function used to train initial network $f_0^{NN}$ remains as MSE loss.

This choice of weighting term increases the importance of accurately fitting regions where the target function exhibits large magnitudes, which in the case of the refinement networks, corresponds to peaks in the target residuals. By emphasizing these high-magnitude regions, the refinement networks are encouraged to reduce errors where they matter most for controlling the $L^{\infty}$ norm. For our purposes, the offset prevents zero-weighting of small-residual points, and the linear form avoids overly concentrating on single outliers. 

We evaluate each function in Table \ref{tab:equations} using a HiPreNet configuration with four stages and network widths [5–10–15–20], employing the tanh activation function and BFGS as the optimizer. For the refinement networks, we compare using the two different loss functions to train them: mean squared error (MSE) and weighted mean squared error (WMSE). The resulting test RMSE and $L^{\infty}$ norm errors are reported in Table \ref{tab:loss_function_metrics_table}. We observe that WMSE can achieve a lower $L^{\infty}$ norm error and even a lower RMSE than standard MSE loss, although the better choice between the two loss functions appears highly problem dependent.
\begin{table}[h!]
    \centering
      \renewcommand{\arraystretch}{1.2}
    \begin{tabular}{cccc}
        \hline
        \textbf{Function} & \textbf{Loss Function} & \textbf{Final RMSE} & \textbf{Final $L^{\infty}$ Norm Error} \\
        \hline
        \multirow{2}{*}{\textbf{I.6.2}}   & MSE & 1.688e-11 & 1.341e-09\\
            & WMSE & 9.520e-12 & 5.298e-10\\
        \hline
        \multirow{2}{*}{\textbf{I.9.18}}  & MSE & 1.535e-07 & 9.822e-05 \\
            & WMSE & 5.377e-07 & 3.732e-05\\
        \hline
         \multirow{2}{*}{\textbf{I.13.12} }& MSE & 2.850e-07 & 1.585e-04\\
         & WMSE & 6.017e-07 & 4.532e-04\\
        \hline
         \multirow{2}{*}{\textbf{I.26.2}}  & MSE & 2.977e-05 & 1.495e-02\\
          & WMSE & 2.028e-05 & 8.061e-03\\
        \hline
         \multirow{2}{*}{\textbf{I.29.16} }& MSE & 1.239e-05 & 1.461e-03\\
           & WMSE & 5.138e-05 & 2.508e-03\\
        \hline
    \end{tabular}
    \caption{Final RMSE and $L^{\infty}$ norm error for each function using MSE and WMSE loss functions in the [5-10-15-20] HiPreNet configuration. Performance differences are highly problem-dependent.}
    \label{tab:loss_function_metrics_table}
\end{table}


In particular, for Function I.29.16 and I.13.12, using WMSE led to worse performance in both RMSE and $L^{\infty}$ norm error compared to standard MSE. This suggests that WMSE may overfit to regions that contain a small subset of data points with large weights. In such cases, the weighting scheme can misdirect learning capacity if residuals exhibit erratic or noisy behavior, causing the model to struggle to generalize. Moreover, as mentioned previously, the $L^{\infty}$ norm itself is inherently challenging to optimize due to its non-differentiability and sensitivity to outliers. WMSE serves as a proxy to mitigate this, but it cannot fully replicate direct $L^{\infty}$ norm minimization.

In summary, while WMSE can be beneficial for reducing worst-case error, it is not a universally superior choice. Selection of loss function should depend on the application's tolerance for average versus maximum error, with WMSE loss potentially being better for one or both metrics depending on the details of the problem.


\subsection{Weighted Sampling Based on Residual Errors}\label{section:weightedsampling}

To further improve the performance of HiPreNet training, we introduce a weighted sampling strategy that leverages information from the current residuals to adaptively guide the training of subsequent networks. Rather than treating all training points equally, we assign greater sampling priority to data points with higher residual error, ensuring that regions with poor approximation are more strongly emphasized in future training. No new data is generated for this approach, it is a resampling of the existing training data. This can also be thought of a data-driven way to choose the weight in the previously described WMSE loss function.

Let the residual at iteration $i$ be defined as
\begin{equation*}
    r_i(x) = f(x) - f^{NN}(x) - \sum_{j=0}^{i-1} e_j \hat{r}^{NN}_j(x)
\end{equation*}
which is then normalized as
\begin{equation*}
    \hat{r}_i(x) = \frac{r_i(x)}{e_i}
\end{equation*}
Given a discrete set of training samples $\textbf{x} = \{x_1,x_2,\dots,x_N\}$, we define a sampling probability $p_k$ at each iteration $i$ for each training sample $x_k \in \textbf{x}$ where
\begin{equation*}
    (p_k)_i = \frac{|\hat{r}_i(x_k)|}{\sum_{j=1}^N |\hat{r}_i(x_j)|}, \quad \text{for } k=1,2,\dots,N
\end{equation*}
to form a discrete probability distribution over the dataset. We use this distribution to resample the training dataset, giving higher selection probability to data points with larger residual error. Each refinement network is now trained on a subset $\left(x_{sampled}, \hat{r}_i(x_{sampled})\right)$ of the original data drawn according to $(p_k)_i$ which incentivizes it to concentrate on high error regions.

This approach is implemented by computing the residuals after each refinement network stage, converting them to sampling probabilities, and drawing a new batch of training samples accordingly. Importantly, this does not increase the overall dataset size, maintaining computational requirements. Algorithm \ref{alg:res2} revises Algorithm \ref{alg:res} to incorporate this method.
\begin{algorithm}
    \caption{HiPreNet Training Algorithm with Weighted Sampling}\label{alg:res2}
    \begin{algorithmic}
        \Require Dataset $(\textbf{x}, \textbf{f}) = \{(x_k, f(x_k))\}^N_{k=1}$, trained neural network $f_{0}^{NN}$, tolerance $tol$
        \State $error \gets L\left(\textbf{f}, f_{0}^{NN}(\textbf{x})\right)$
        \State $r_0 \gets \textbf{f} - f_{0}^{NN}(\textbf{x})$
        \State $e_0 \gets \max{|r_0(\textbf{x})|}$ \Comment{estimate of true $e_0$}
        \State $\Hat{r}_0(\textbf{x}) \gets \frac{r_0}{e_0}$
        \State $p_{0}(\textbf{x}) \gets \frac{|\hat{r}_{0}(\textbf{x})|}{\sum_{j=1}^N |\hat{r}_{0}(x_j)|}$
        \State $\left(\textbf{x}_{sampled}, \Hat{r}_{0}(\textbf{x}_{sampled})\right) \gets sample\left(p_0(\textbf{x})\right)$
        \State $\Hat{r}^{NN}_{0}(\textbf{x}) \gets train(\textbf{x}_{sampled}, \Hat{r}_{0}(\textbf{x}_{sampled}))$
        \State $i \gets 1$
        \While{$error > tol$}
            \State  $f^{NN}_{i}(\textbf{x}) \gets f_{i-1}^{NN} + e_{i-1} \Hat{r}_{i-1}^{NN}(\textbf{x})$
            \State $r_{i}(\textbf{x}) \gets \textbf{f} - f^{NN}_{i}(\textbf{x})$
            \State $e_{i} \gets \max{|r_i(\textbf{x})|}$ \Comment{estimate of true $e_{i}$}
            \State $\Hat{r}_{i}(\textbf{x}) \gets \frac{r_{i}(\textbf{x})}{e_{i}}$
            \State $p_{i}(\textbf{x}) \gets \frac{|\hat{r}_{i}(\textbf{x})|}{\sum_{j=1}^N |\hat{r}_{i}(x_j)|}$
            \State $\left(\textbf{x}_{sampled}, \Hat{r}_{i}(\textbf{x}_{sampled})\right) \gets sample\left(p_{i}(\textbf{x})\right)$
            \State $\Hat{r}^{NN}_{i}(\textbf{x}) \gets train(\textbf{x}_{sampled}, \Hat{r}_{i}(\textbf{x}_{sampled}))$
            \State $error \gets L\left(\textbf{f}, f_{i}^{NN}(\textbf{x})\right)$
            \State $i \gets i + 1$
        \EndWhile
    \end{algorithmic}
\end{algorithm}

To evaluate the method, we applied weighted sampling across all five benchmark functions from the Feynman dataset using four-stage HiPreNets with widths [5–10–15–20], choosing the lowest error across three simulations with different random seeds. A weighted resampling of the dataset before training each refinement networks focuses them on handling more challenging regions of the dataset without increasing dataset size, offering a computationally efficient route to reducing error. The sampling strategy generally maintains good overall coverage of the domain while prioritizing regions with higher residual error. In particular, samples near the boundaries of the domain, where the residuals tend to be largest, are selected more frequently during training. This suggests that the weighted sampling method effectively concentrates training effort on the areas that are most difficult to approximate, without requiring an increase in dataset size.

\begin{figure}
    \centering
    \includegraphics[width=0.65\linewidth]{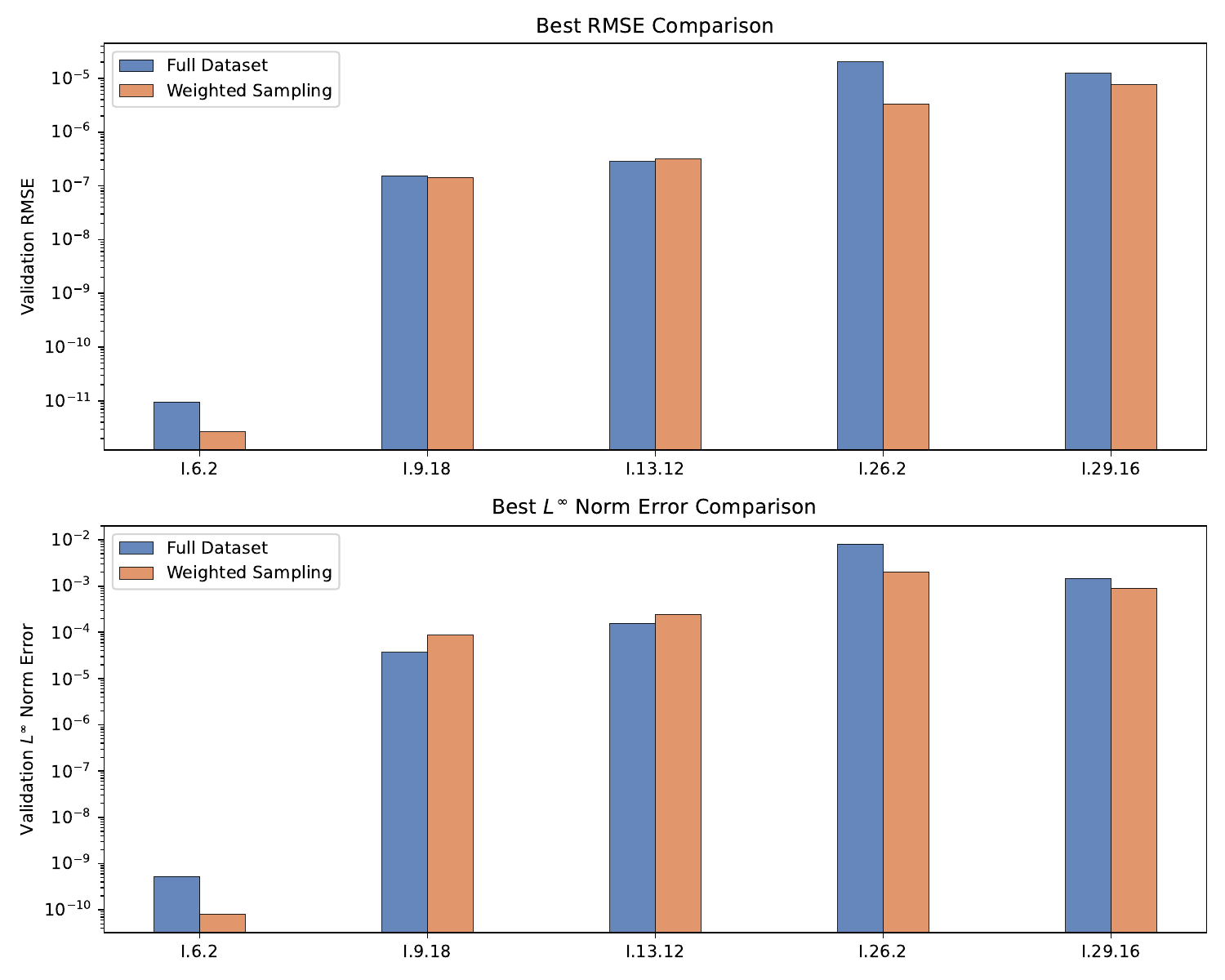}
    \caption{Comparison of best-achieved validation RMSE and $L^{\infty}$ norm error across all five functions, with and without weighted sampling. For each function, the better of MSE and WMSE loss is shown. Weighted sampling reduces both error metrics for four of the five functions.}
    \label{fig:weightedbar}
\end{figure}

The final validation RMSE and $L_\infty$ norm error for each function are given in Table \ref{tab:weighted_sampling_metrics_table}, where we see that using weighted sampling with MSE loss gives the lowest RMSE and $L^{\infty}$ norm error for all functions except for I.29.16. The best results for each function from Tables \ref{tab:loss_function_metrics_table} where no weighted sampling was used and \ref{tab:weighted_sampling_metrics_table} where weighted sampling was used are compared in bar charts in Figure \ref{fig:weightedbar}. We see that for four out of the five functions, weighted sampling successfully reduces both types of error. 

To summarize, this method can lead to a noticeable reduction in both RMSE and $L^{\infty}$ norm error. Similar to what was observed for the weighted mean square error loss function, weighted sampling does not guarantee a reduction in either error metric. It may cause the models to overfit to the training data, reducing their ability to generalize and worsening their performance on the validation data. Additionally, the weighted sampling technique is one instance of a broader principle: identify regions where the current model performs poorly and concentrate training resources there. In settings where new data can be generated cheaply, this principle can alternatively be implemented by augmenting the training dataset with new samples drawn from high error regions, rather than resampling existing data. We return to this idea in Section \ref{section:highdim} in the context of a high dimensional power systems example, where the simulator allows cheap data generation and the high error region has a clear physical interpretation.
\begin{table}[h!]
    \centering
      \renewcommand{\arraystretch}{1.2}
    \begin{tabular}{cccc}
        \hline
        \textbf{Function} & \textbf{Loss} & \textbf{Final RMSE} & \textbf{Final $L^{\infty}$ Norm Error} \\
        \hline
        \multirow{2}{*}{\textbf{I.6.2}} & MSE & 2.675e-12  & 8.111e-11 \\
            & WMSE & 1.352e-11  & 3.830e-10 \\
        \hline
        \multirow{2}{*}{\textbf{I.9.18}} & MSE & 1.429e-07  & 8.714e-05 \\
         & WMSE & 1.730e-07  & 1.201e-04 \\
        \hline
        \multirow{2}{*}{\textbf{I.13.12}} & MSE & 3.169e-07  & 2.473e-04 \\
         & WMSE & 5.824e-07  & 4.729e-04 \\
        \hline
        \multirow{2}{*}{\textbf{I.26.2}} & MSE & 3.376e-06  & 2.018e-03 \\
         & WMSE & 7.085e-06  & 3.900e-03 \\
        \hline
        \multirow{2}{*}{\textbf{I.29.16} }& MSE & 1.544e-05  & 3.555e-03 \\
        & WMSE & 7.640e-06  & 8.989e-04 \\
        \hline
    \end{tabular}
    \caption{The final validation RMSE and $L^{\infty}$ norm error for both MSE and WMSE loss functions using the weighted sampling technique. Using weighted sampling with MSE loss results in lower error than using weighted sampling with WMSE loss, aside from for Function I.29.16.}
    \label{tab:weighted_sampling_metrics_table}
\end{table}

\subsection{Targeted Model Correction with Local Residual Patching}\label{section:patching}
\begin{figure}[h]
    \centering
    \includegraphics[width=0.85\textwidth]{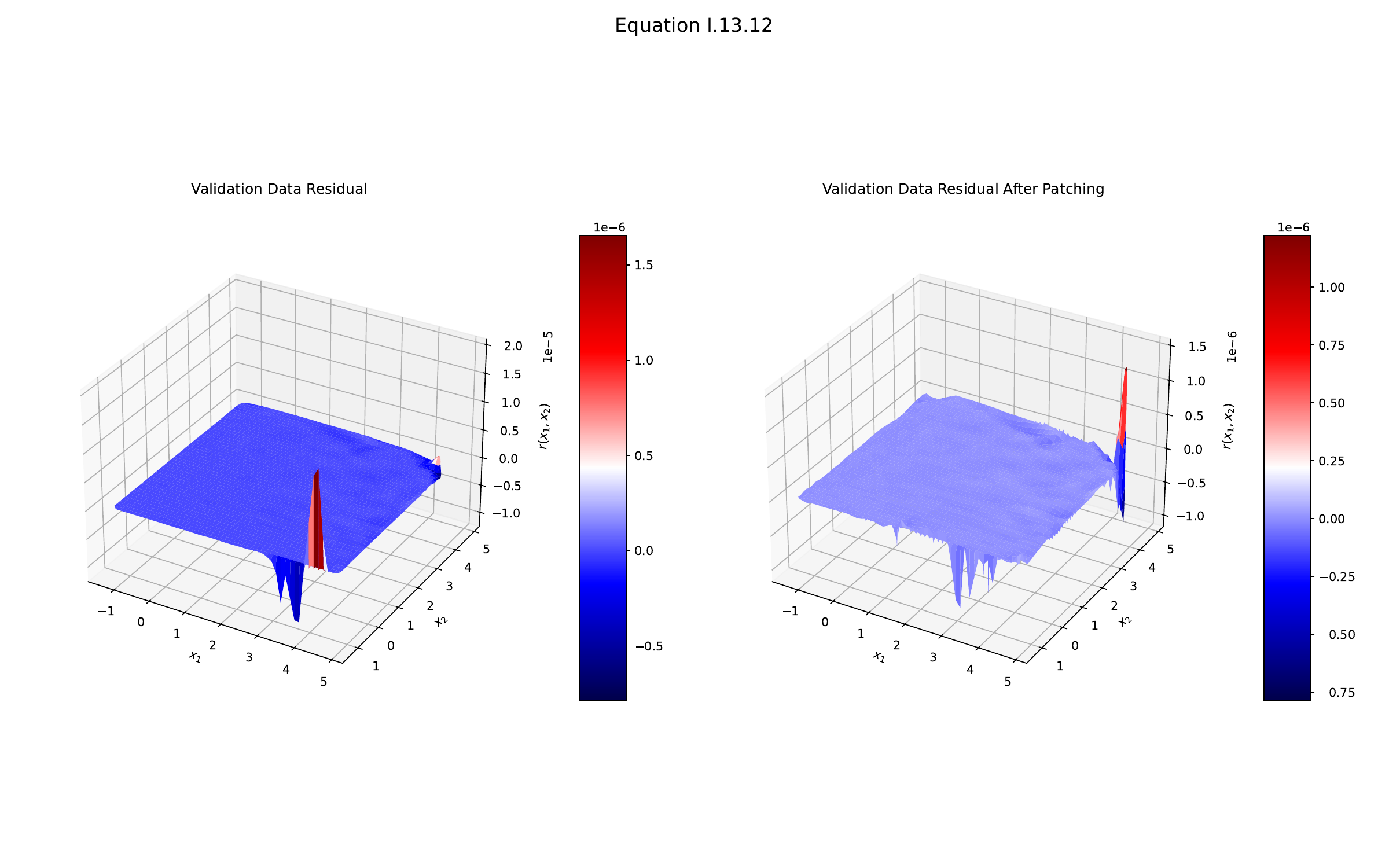}
    \caption{Validation data residuals for Function I.13.12: the left plot shows final residuals after standard HiPreNet training and the right plot shows residuals after applying the localized patch network. The spike near the domain boundary, corresponding to the $L^\infty$ norm error, is reduced by nearly an order of magnitude Original Patching. Note that axis and colorbar scales differ across plots. The spike in the right plot is the location of the new $L^\infty$ norm error.}
    \label{fig:val_patching}
\end{figure}

For a given problem, even after the HiPreNet model has been fully trained, a dominant spike in the residuals may remain in a localized region of the input domain. This is apparent in Figure \ref{fig:val_patching} (left), where the final residuals for a model trained to approximate Function I.13.12 exhibit a large peak on the boundary, corresponding to the location of the $L^{\infty}$ norm error. This observation motivates a targeted post-training correction, where rather than retraining the entire ensemble, we train a small ``patching" network focused exclusively on the neighborhood of the spike.

In this approach, once the HiPreNet model is trained after $m$ iterations, we evaluate its performance on a validation dataset to ``probe" the model's weaknesses. With $r_m(x) = f(x)-f_m^{NN}(x)$ as the final residuals of the model, where $f(x)$ is the true target and $f_m^{NN}(x)$ is the model after $m$ iterations, we identify the validation point $x_{max}=\arg \max |r_m(x)|$ that has the maximum absolute residual. We next define a Gaussian weight function centered on $x_{max}$
\begin{equation*}
    w(x) = \exp\left(-\frac{\|x-x_{max}\|_2^2}{2\sigma^2}\right) = \exp\left(-\frac{d^2}{2\sigma^2}\right)
\end{equation*}
where $d = \|x-x_{max}\|_2$ is the distance from a given point $x$ to $x_{max}$ and $\sigma$ controls the extent of the neighborhood around the spike. We then look at only the neighborhood $N(x_{max}) = \{x \in \text{validation dataset} \ | \ w(x) > \tau \}$ of the $K = |N(x_{max})|$ validation points around $x_{max}$ that satisfy $w(x) > \tau$ for a set threshold $\tau$. With this weight function, a patching network $f_{patch}(x)$ is trained on this neighborhood of points by using the loss function
\begin{equation*}
    L_{patch} = \frac{1}{K}\sum_{i=1}^K w(x_i)\left(\hat{r}_m(x_i) - f_{patch}(x_i)\right)^2, \quad x_i\in N(x_{max})
\end{equation*}
where $\hat{r}_m(x_i) = r_m(x_i)/\max(|r_m|)$ are the normalized final residuals. Once this network is trained, the final corrected prediction is
\begin{equation*}
    f_m^{patch}(x) = f_m^{NN}(x) + \mathds{1}(x) \max(|r_m|) f_{patch}(x)
\end{equation*}
where the indicator function
\begin{equation*}
    \mathds{1}(x) = \begin{cases}
        1 \ \text{if} \ x \in N(x_{max})\\
        0 \ \text{otherwise}
    \end{cases}
\end{equation*}
ensures that the patching network is applied only to inputs within the patching neighborhood $N(x_{max})$, leaving all other inputs unchanged.

The patching network $f_{patch}$ consists of two hidden layers of 16 neurons each with tanh activation functions, giving a total of 337 trainable parameters. The network is optimized using the BFGS algorithm with a gradient tolerance of $10^{-12}$ and a maximum of 20,000 iterations, consistent with the optimizer used for the HiPreNet refinement networks. Importantly, this network also uses a tanh activation applied to the output, constraining its predictions to $[-1, 1]$ for any input. This is a critical design choice with two important consequences. First, since the normalized training targets lie in $[-1, 1]$ by construction, the tanh does not restrict the network's expressivity within its training distribution. Second, it provides a hard guarantee that the correction magnitude never exceeds $\max|r_m|$ for any input, including unseen test points. Without this constraint, we observed the network extrapolating to corrections orders of magnitude larger than intended on slightly different test inputs, catastrophically worsening predictions. We proceed to use this setup to test all of the following examples.

For this problem, we use $\sigma = 0.5$ and $\tau = 10^{-3}$, retaining approximately 21,000 of 1,000,000 validation points for training. The threshold $\tau = 10^{-3}$ was chosen as a practical cutoff for computational efficiency. For any choice of $\sigma$, a point at distance $d$ from $x_{\max}$ is retained if 
\begin{equation*}
    \exp(-d^2 / 2\sigma^2) > \tau,
\end{equation*}
which corresponds to
\begin{equation*}
    d \leq \sigma\sqrt{-2\ln\tau}.
\end{equation*}
This means setting $\tau = 10^{-3}$ retains all points within approximately $3.7\sigma$ of $x_{\max}$. As such, points with $w(x) < \tau$ contribute less than 0.1\% of the maximum weight to the loss function, and their inclusion would increase training cost without meaningfully affecting the learned correction. 

The parameter $\sigma$ was selected by sweeping over $\{0.10, 0.25, 0.50, 0.75, 1.00, 1.50\}$ and examining the sensitivity of the results, shown in Table \ref{tab:sigma_sweep}. Values of $\sigma$ below 0.25 retained too few points for reliable training and showed no improvement in validation $L^\infty$ norm error. The value of $\sigma = 0.75$ was chosen as a representative value as it is the smallest value that retains enough points $n$ for reliable training and attains the lowest observed validation $L^{\infty}$ norm error.

\begin{table}[h]
\centering
\begin{tabular}{crcc}
\hline
\textbf{$\sigma$} & \textbf{$n$} & \textbf{Val RMSE} & \textbf{Val $L^\infty$ Norm Error} \\
\hline
0.10 & 378     & 1.531e-08 & 1.356e-05 \\
0.25 & 3,634   & 1.460e-08 & 1.384e-05 \\
0.50 & 21,410  & 4.102e-09 & 1.726e-06 \\
0.75 & 64,768  & 4.824e-09 & 1.559e-06 \\
1.00 & 164,519 & 3.541e-09 & 1.559e-06 \\
1.50 & 906,250 & 3.786e-09 & 1.559e-06 \\
\hline
\end{tabular}
\caption{Sensitivity of patching results to the parameter $\sigma$, showing validation RMSE and $L^\infty$ norm error alongside the number of training points retained above threshold $\tau = 10^{-3}$. Based on these validation results, $\sigma = 0.75$ was selected as it is the smallest value in the stable range that retains enough points for reliable training. Values of $\sigma$ below 0.25 retain too few points for reliable training.}
\label{tab:sigma_sweep}
\end{table}

\begin{figure}[h]
    \centering
    \includegraphics[width=0.85\textwidth]{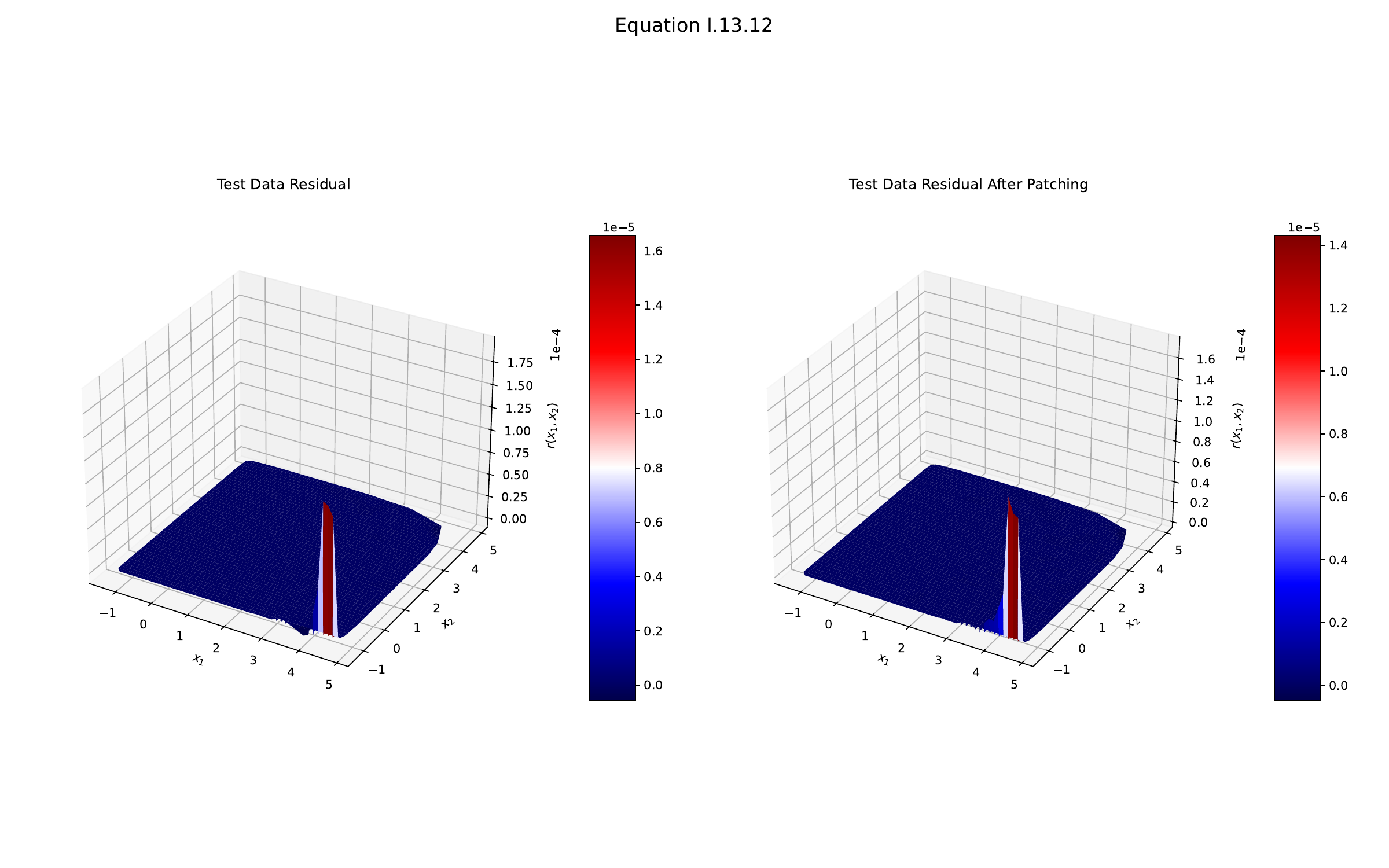}
    \caption{Test data residuals for Function I.13.12: the left plot shows final residuals after standard HiPreNet training and the right plot shows residuals after applying the localized patch network trained on validation data. The spike is partially reduced, with the $L^\infty$ norm error decreasing by approximately 11\%. The smaller improvement relative to validation reflects the test spike being approximately $9\times$ larger in magnitude than the validation spike, which limits the correction achievable within the bounded output of the patching network. Note that axis and colorbar scales differ across plots.}
    \label{fig:test_patching}
\end{figure}

We evaluate the selected configuration on an independently generated test dataset of 1,000,000 samples drawn from the same distribution as the validation data but with a different random seed, ensuring no overlap with the data used to train the patching network or select $\sigma$. The results are shown in Table \ref{tab:patching_results} and illustrated in Figures \ref{fig:val_patching} and \ref{fig:test_patching}. On the validation data, patching reduces RMSE from 2.758e-08 to 4.824e-09 and $L^\infty$ norm error from 2.293e-05 to 1.559e-06, which are approximately 83\% and 93\% decreases, respectively. This is visible in Figure \ref{fig:val_patching} as a dramatic reduction in the residual spike. On the independently generated test data, the $L^\infty$ norm error is reduced from 2.127e-04 to 1.898e-04, a reduction of approximately 11\%, with RMSE slightly decreasing from 2.271e-07 to 2.001e-07.

\begin{table}[h]
\centering
\begin{tabular}{llcc}
\hline
\textbf{Data} & \textbf{Model} & \textbf{RMSE} & \textbf{$L^\infty$ Norm Error} \\
\hline
\multirow{2}{*}{Validation} & HiPreNet Stage 4 & 2.758e-08 & 2.293e-05 \\
                             & Original Patching   & 4.824e-09 & 1.559e-06 \\
\hline
\multirow{2}{*}{Test}       & HiPreNet Stage 4 & 2.271e-07 & 2.127e-04 \\
                             & Original Patching   & 2.001e-07 & 1.898e-04 \\
\hline
\end{tabular}
\caption{RMSE and $L^\infty$ norm error for Function I.13.12 before and after local residual patching, evaluated on validation and test data.}
\label{tab:patching_results}
\end{table}

The smaller improvement on test data reflects an inherent limitation of the technique. The $L^{\infty}$ norm error point on the test data was found to lie within the patch region, with a distance of $0.27$ from the $L^{\infty}$ norm error point on the validation data and a Gaussian weight of $w(x_{\max}^{test}) = \exp(-0.27^2 / (2 \times 0.75^2)) \approx 0.93 > \tau = 10^{-3}$. Even though the test spike is inside the patching region, applying the tanh activation to the network's outputs guarantees that the correction magnitude never exceeds the magnitude of the validation spike, which is $\max|r_m| = 2.293\text{e-05}$. The test spike is approximately $9\times$ larger at 2.127e-04, so the patch can only partially correct it. This indicates that the validation dataset is not large or robust enough to properly capture the spike, which suggests a method for targeted sampling if further data generation is allowed, which we investigate in subsection \ref{targetedsampling}.

Returning to the assumption that no more data can be generated for training, we apply the patching technique to the HiPreNet models trained for each of the Feynman functions, and report the results in Table \ref{tab:patching_all}. For each function aside from Function I.29.16, we choose the $\sigma$ that retains the largest amount of samples under 100000 samples to ensure sufficient coverage of the patching region without introducing overfitting. For Function I.29.16, we choose $\sigma = 0.75$ to demonstrate a potential failure case for the method. After applying patching, there is a drop in the test $L^{\infty}$ norm error for all functions aside from Function I.29.16, where it moderately increases. Of note, the test $L^{\infty}$ norm error for function I.6.2 is reduced by approximately 59\%, and the test $L^{\infty}$ norm error for function I.26.2 is reduced by approximately 52\%.

\begin{figure}[h]
    \centering
    \includegraphics[width=0.9\textwidth]{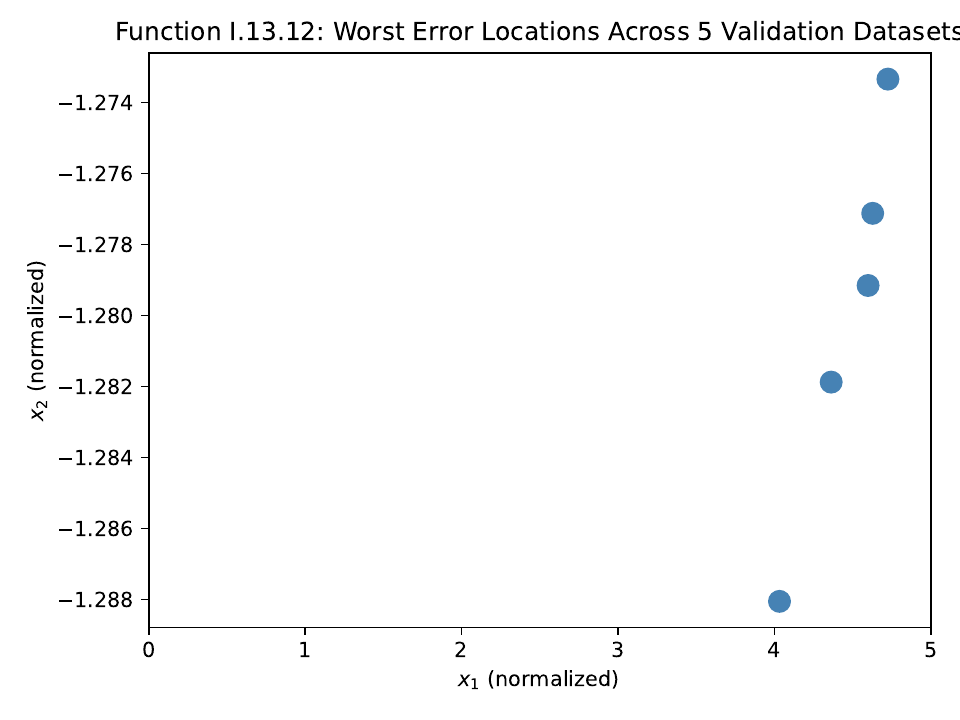}
    \caption{$L^{\infty}$ norm error locations across five independently generated validation datasets for Function I.13.12, plotted in the normalized input space. The locations cluster tightly near the same boundary region across all five datasets, indicating that the $L^{\infty}$ norm error location is stable and the patch network trained on one dataset will be applied in the correct region for unseen data.}
    \label{fig:worsterror_patching_I1312}
\end{figure}

\begin{figure}[h]
    \centering
    \includegraphics[width=0.9\textwidth]{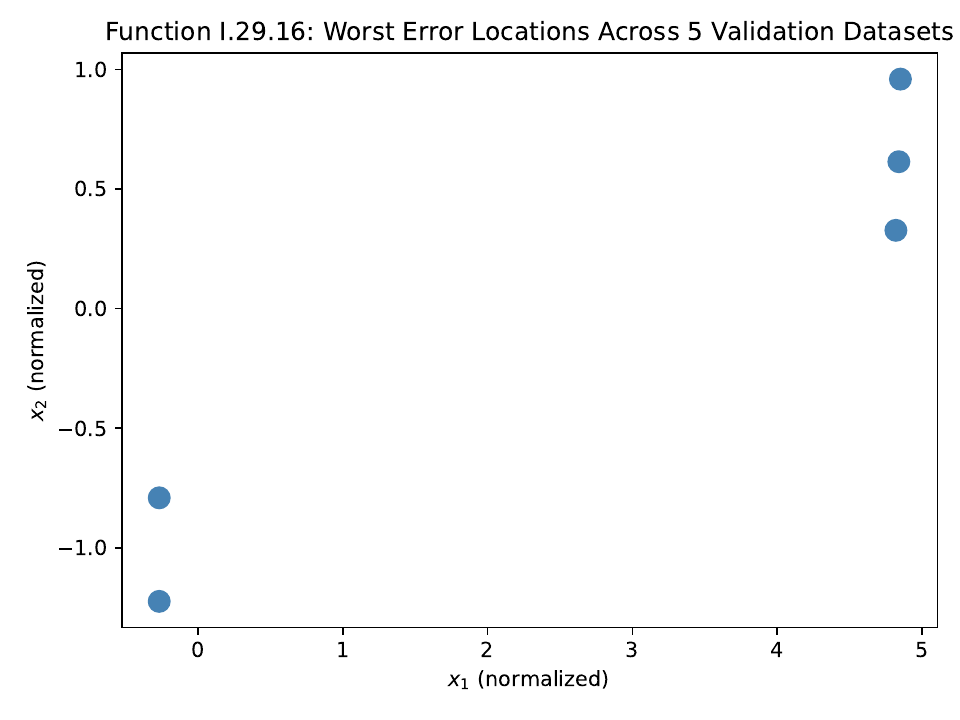}
    \caption{$L^{\infty}$ norm error locations across five independently generated validation datasets for Function I.29.16, plotted in the $x_1-x_2$ plane of the normalized input space. The locations fall into two distinct clusters separated by a distance of approximately 5.3, corresponding to two different boundary corners of the domain. When the validation $L^{\infty}$ norm error falls in one cluster and the test $L^{\infty}$ norm error falls in the other, the patch network is applied in the wrong region and cannot improve performance.}
    \label{fig:worsterror_patching_I2916}
\end{figure}

To understand why patching worsens performance on the test data for Function I.29.16, we examine the stability of the $L^{\infty}$ norm error location across independently generated validation datasets. Figures \ref{fig:worsterror_patching_I1312} and \ref{fig:worsterror_patching_I2916} show the $L^{\infty}$ norm error locations across five validation datasets in the normalized input space for Function I.13.12 and I.29.16, respectively. Unlike Function I.13.12, where the $L^{\infty}$ norm error point was consistently located near the same boundary region, Function I.29.16 exhibits two distinct clusters of $L^{\infty}$ norm error locations separated by a distance of approximately 5.3 in the normalized input space. The patch network is trained on whichever cluster appears in the validation data, but has no effect if the test worst error falls in the other cluster. As such, when the worst error location is not stable, the technique cannot reliably generalize to unseen test data using only one patching network, suggesting that an iterative training process to construct multiple patching networks is necessary.

\subsubsection{Iterative Patching}
To test this idea, we apply the iterative patching procedure to Function I.29.16. Beginning from the HiPreNet Stage 4 predictions, we identify the largest residual sample on the validation data and train a patching network as described above. We then update the running predictions on both validation and test data, and repeat the process, meaning at each subsequent iteration, the largest residual sample is re-identified from the updated residuals, and a new patching network is trained and applied. This continues for a fixed number of iterations $P$, resulting in a stack of patching networks $\{f_{patch}^{(p)}\}_{p=1}^{P}$ with associated scales $\{\max|r^{(p)}|\}$ and largest residual samples $\{x_{max}^{(p)}\}$. Then, the corrected prediction at iteration $p$ is

\begin{equation*}
    f^{(p)}(x) = f^{(p-1)}(x) + \mathds{1}^{(p)}(x)\max\left|r^{(p-1)}\right| f_{patch}^{(p)}(x),
\end{equation*}

where $r^{(p-1)}(x) = f(x) - f^{(p-1)}(x)$ is the residual after $p-1$ patches, $f^{(0)} = f_m^{NN}$ is the Stage 4 HiPreNet output, and $\mathds{1}^{(p)}$ is the indicator function for the $p$-th patch neighborhood $N(x_{max}^{(p)})$.

By construction, each patch network targets the current dominant spike in the residuals, and successive iterations address distinct high error regions as earlier ones are corrected. This is the behavior needed for Function I.29.16, where the $L^\infty$ norm error location is unstable across two different regions. If the first patch corrects the spike in one region, the second iteration's largest error search will naturally migrate to the other region, and a second patch network can be trained and applied there.

The results of applying $P = 2$ iterative patches to Function I.29.16 are shown in Table \ref{tab:iterative_patching_I2916}. On the validation data, the $L^\infty$ norm error decreases monotonically across iterations, with the largest reduction occurring in the first two patches as the two dominant error clusters are successively corrected. On the test data, which was not used at any stage of the patching procedure, the $L^\infty$ norm error also decreases across iterations, confirming that the technique generalizes beyond the validation set when the iterative procedure is used. The improvement on test data is more modest than on validation, consistent with the bounded-output guarantee limiting the magnitude of each correction to at most $\max|r^{(p-1)}|$ on the validation residuals, which may underestimate the true spike magnitude on the test data.
\begin{table}[h]
\centering
\begin{tabular}{llcc}
\hline
\textbf{Data} & \textbf{Model} & \textbf{RMSE} & \textbf{$L^\infty$ Norm Error} \\
\hline
\multirow{6}{*}{Validation} & Initial         & 6.4418e-04 & 3.1251e-02 \\
                             & Stage 2         & 8.6174e-05 & 8.9618e-03 \\
                             & Stage 3         & 3.4741e-05 & 3.4392e-03 \\
                             & Stage 4         & 1.2301e-05 & 1.2590e-03 \\
                             & Patch 1         & 1.2107e-05 & 9.4403e-04 \\
                             & Patch 2         & 1.1742e-05 & 9.3998e-04 \\
\hline
\multirow{6}{*}{Test}       & Initial         & 6.4112e-04 & 3.1083e-02 \\
                             & Stage 2         & 8.5425e-05 & 9.0803e-03 \\
                             & Stage 3         & 3.4677e-05 & 2.8813e-03 \\
                             & Stage 4         & 1.2255e-05 & 9.5549e-04 \\
                             & Original Patching  & 1.1777e-05 & 8.9009e-04 \\
\hline
\end{tabular}
\caption{RMSE and $L^\infty$ norm error for Function I.29.16 across HiPreNet ensemble stages and iterative Gaussian patching, evaluated on validation and test data. Patch iterations are applied sequentially to the Stage 4 predictions; intermediate test results are not reported as the patch networks are selected using validation data only.}
\label{tab:iterative_patching_I2916}
\end{table}

We note that the iterative procedure introduces no additional hyperparameters beyond the number of iterations $P$, a decay rate $\gamma \in (0,1]$, and the shared $\tau$ used in each step. At iteration $p$, the neighborhood width is set to $\sigma^{(p)} = \sigma_0 \cdot \gamma^{p}$, where $\sigma_0$ is the initial value selected by the sweep described above. This decay ensures that successive patch networks are trained on progressively smaller neighborhoods, concentrating each correction more tightly around the newly identified worst-error point as the dominant spikes are resolved. The total computational cost scales linearly with $P$, since each iteration trains one additional patch network of fixed size using BFGS.

\begin{table}[h!]
\centering
\begin{tabular}{lcccccc}
\hline
\textbf{Function} & \textbf{$\sigma$} & \textbf{Data} & \textbf{Model} & \textbf{RMSE} & \textbf{$L^\infty$ Norm Error} \\
\hline
\multirow{4}{*}{\shortstack[l]{I.6.2\\(2D)}} & \multirow{4}{*}{0.25}
    & \multirow{2}{*}{Validation} & HiPreNet Stage 4 & 1.678e-11 & 1.342e-09 \\
    &                             &                  & Original Patching   & 1.502e-11 & 2.930e-10 \\
    &                             & \multirow{2}{*}{Test}       & HiPreNet Stage 4 & 1.698e-11 & 1.551e-09 \\
    &                             &                  & Original Patching   & 1.507e-11 & 6.367e-10 \\
\hline
\multirow{4}{*}{\shortstack[l]{I.9.18\\(6D)}} & \multirow{4}{*}{0.75}
    & \multirow{2}{*}{Validation} & HiPreNet Stage 4 & 1.363e-07 & 4.388e-05 \\
    &                             &                  & Original Patching   & 1.175e-07 & 3.833e-05 \\
    &                             & \multirow{2}{*}{Test}       & HiPreNet Stage 4 & 2.364e-07 & 1.831e-04 \\
    &                             &                  & Original Patching   & 2.202e-07 & 1.393e-04 \\
\hline
\multirow{4}{*}{\shortstack[l]{I.13.12\\(2D)}} & \multirow{4}{*}{0.75}
    & \multirow{2}{*}{Validation} & HiPreNet Stage 4 & 2.758e-08 & 2.293e-05 \\
    &                             &                  & Original Patching   & 4.824e-09 & 1.559e-06 \\
    &                             & \multirow{2}{*}{Test}       & HiPreNet Stage 4 & 2.271e-07 & 2.127e-04 \\
    &                             &                  & Original Patching   & 2.001e-07 & 1.898e-04 \\
\hline
\multirow{4}{*}{\shortstack[l]{I.26.2\\(2D)}} & \multirow{4}{*}{0.25}
    & \multirow{2}{*}{Validation} & HiPreNet Stage 4 & 2.431e-05 & 8.549e-03 \\
    &                             &                  & Original Patching   & 1.357e-05 & 3.578e-03 \\
    &                             & \multirow{2}{*}{Test}       & HiPreNet Stage 4 & 2.977e-05 & 1.495e-02 \\
    &                             &                  & Original Patching   & 1.600e-05 & 7.167e-03 \\
\hline
\multirow{4}{*}{\shortstack[l]{I.29.16\\(3D)}} & \multirow{4}{*}{0.75}
    & \multirow{2}{*}{Validation} & HiPreNet Stage 4 & 1.230e-05 & 1.259e-03 \\
    &                             &                  & Original Patching   & 1.209e-05 & 9.440e-04 \\
    &                             & \multirow{2}{*}{Test}       & HiPreNet Stage 4 & 1.226e-05 & 9.555e-04 \\
    &                             &                  & Original Patching   & 1.217e-05 & 1.398e-03 \\
\hline
\end{tabular}
\caption{RMSE and $L^\infty$ norm error before and after local residual patching for all five Feynman functions, evaluated on validation and test data. Function dimensionality is left-aligned below the function name.}
\label{tab:patching_all}
\end{table}

\subsubsection{Targeted Sampling}\label{targetedsampling}
As mentioned previously, the limited gains in accuracy on the test dataset observed with Function I.13.12 appear to be a result of the validation dataset not being enough to completely capture the spike in the residuals. In other words, the small area selected for patching is generally noisier and more oscillatory than the whole large scale original dataset. This suggests that there needs to be a dense dataset in the patching area to allow the model to completely capture the spike. With the ability to generate more data, this prompts us to generate a new dataset comprised of data that falls within a selected patching area, and then train the patching network on this dataset.

In practice, we calculate $x_{max}$ and the neighborhood around it, $N(x_{max})$, as previously discussed. We then generate a dataset of 50000 samples that fall within the domain encapsulated by $N(x_{max})$ and train the patching network on this dataset. We compare the performance of this approach to the previous approach in Table \ref{tab:patching_targeted} on the same datasets, where we observe that using this targeted sampling approach consistently leads to lower test $L^{\infty}$ norm error, aside from Function I.29.16, whose degradation in performance we discussed previously.
\begin{table}[h!]
\centering
\begin{tabular}{lccccc}
\hline
\textbf{Function} & \textbf{$\sigma$} & \textbf{Data} & \textbf{Model} & \textbf{RMSE} & \textbf{$L^\infty$ Norm Error} \\
\hline
\multirow{4}{*}{\shortstack[l]{I.6.2\\(2D)}} & \multirow{4}{*}{0.25}
    & \multirow{2}{*}{Validation} & Original Patching      & 1.502e-11 & 2.930e-10 \\
    &                             &                     & Targeted Patching & 1.529e-11 & 2.930e-10 \\
    &                             & \multirow{2}{*}{Test}       & Original Patching      & 1.507e-11 & 6.367e-10 \\
    &                             &                     & Targeted Patching & 1.526e-11 & 2.946e-10 \\
\hline
\multirow{4}{*}{\shortstack[l]{I.9.18\\(6D)}} & \multirow{4}{*}{0.75}
    & \multirow{2}{*}{Validation} & Original Patching      & 1.175e-07 & 3.833e-05 \\
    &                             &                     & Targeted Patching & 1.206e-07 & 3.833e-05 \\
    &                             & \multirow{2}{*}{Test}       & Original Patching      & 2.202e-07 & 1.393e-04 \\
    &                             &                     & Targeted Patching & 1.793e-07 & 9.822e-05 \\
\hline
\multirow{4}{*}{\shortstack[l]{I.13.12\\(2D)}} & \multirow{4}{*}{0.75}
    & \multirow{2}{*}{Validation} & Original Patching      & 4.824e-09 & 1.559e-06 \\
    &                             &                     & Targeted Patching & 1.346e-08 & 8.670e-06 \\
    &                             & \multirow{2}{*}{Test}       & Original Patching      & 2.001e-07 & 1.898e-04 \\
    &                             &                     & Targeted Patching & 1.110e-07 & 8.867e-05 \\
\hline
\multirow{4}{*}{\shortstack[l]{I.26.2\\(2D)}} & \multirow{4}{*}{0.25}
    & \multirow{2}{*}{Validation} & Original Patching      & 1.357e-05 & 3.578e-03 \\
    &                             &                     & Targeted Patching & 1.325e-05 & 3.578e-03 \\
    &                             & \multirow{2}{*}{Test}       & Original Patching      & 1.600e-05 & 7.167e-03 \\
    &                             &                     & Targeted Patching & 1.508e-05 & 6.261e-03 \\
\hline
\multirow{4}{*}{\shortstack[l]{I.29.16\\(3D)}} & \multirow{4}{*}{0.75}
    & \multirow{2}{*}{Validation} & Original Patching      & 1.209e-05 & 9.440e-04 \\
    &                             &                     & Targeted Patching & 1.209e-05 & 9.440e-04 \\
    &                             & \multirow{2}{*}{Test}       & Original Patching      & 1.217e-05 & 1.398e-03 \\
    &                             &                     & Targeted Patching & 1.217e-05 & 1.398e-03 \\
\hline
\end{tabular}
\caption{RMSE and $L^\infty$ norm error comparing standard and augmented local residual patching for all five Feynman functions, evaluated on validation and test data. Function dimensionality is left-aligned below the function name.}
\label{tab:patching_targeted}
\end{table}

\section{Improving Boundary Accuracy through Domain Expansion}\label{section:domainexpand}
The largest errors in training neural network models often occur along the boundary of the training domain. To address this issue, we propose training the neural network on an input domain that is larger than the domain of interest. This technique allows the neural network to better approximate the boundary in the true evaluation domain, reducing the overall RMSE and $L^{\infty}$ norm error. In practice, this technique could be used for applications such as material science and manufacturing, where training data can be collected for features such as temperature and strain that is outside the expected domain in operation, but can still be collected in practice.

Let $f(x):D  \to R$ be a continuous function defined on compact domain $D \subset R^n$. We approximate $f$ with a neural network $f^{NN}(x)$ by training it on a larger domain $\hat{D} \supset D$. This approach aims to mitigate boundary-related errors by allowing the network to learn beyond the boundary, thereby improving its robustness and performance on the target domain $D$.

\begin{figure}
    \centering
    \includegraphics[width=0.7\linewidth]{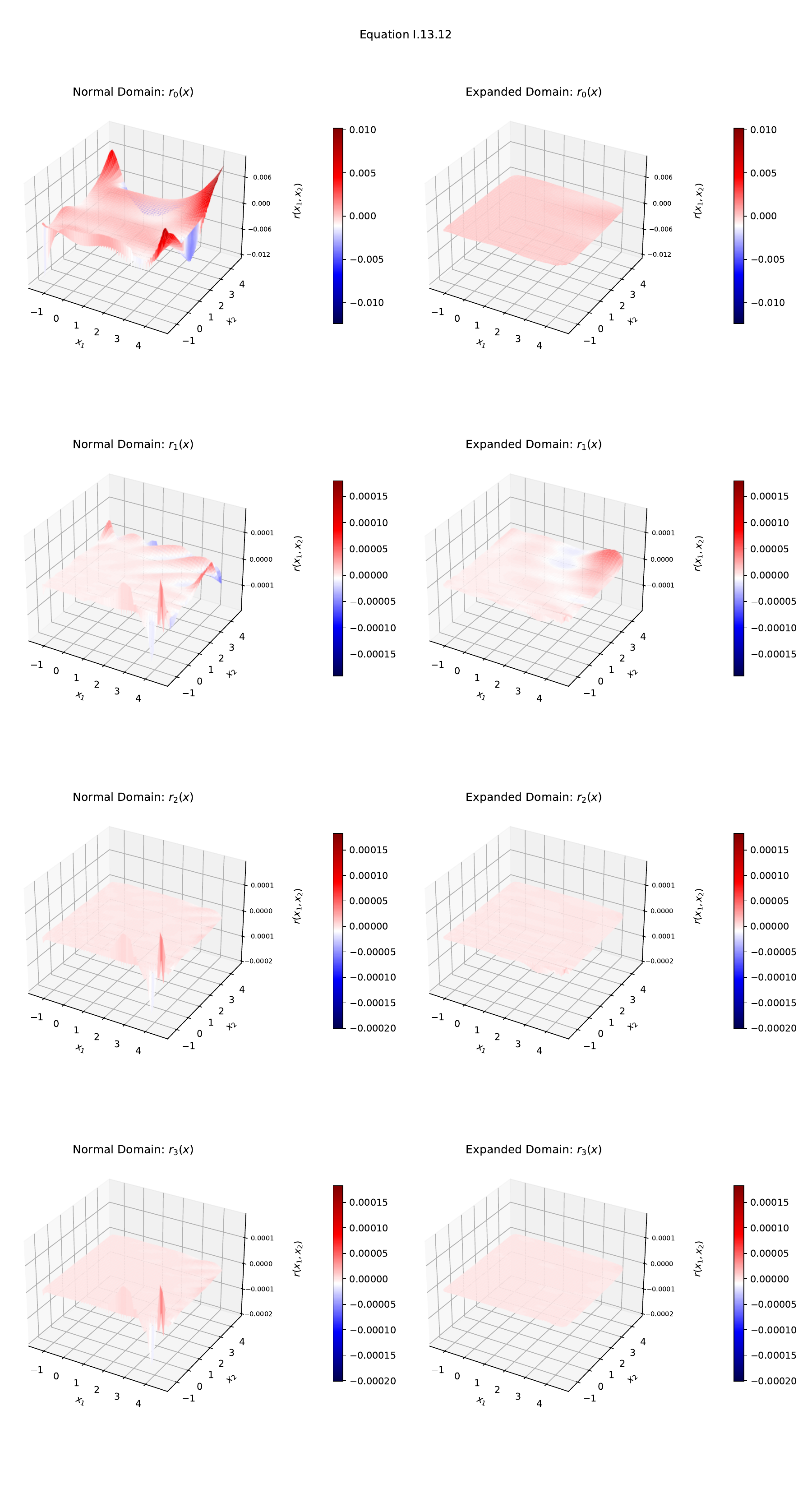}
    \caption{Residuals at each HiPreNet stage for Function I.13.12, evaluated on validation data in $[1.1, 4.9]$. The standard network (left column) is trained on the same domain, while the expanded network (right column) is trained on the larger domain $[1, 5]$. The expanded training domain produces smaller, more uniform residuals by reducing boundary errors.}
    \label{fig:expanded_domain}
\end{figure}

We evaluate this approach on Function I.13.12, where 1,000,000 validation samples are generated with all input variables in $[1.1,4.9]$. Two residual ensembles are trained: one on 1,000,000 samples from the domain $[1.1, 4.9]$ (the baseline), and another on the slightly expanded domain $[1, 5]$. We compare each ensemble's performance on the validation data in Table \ref{tab:expanded_boundary_table} and in Figure \ref{fig:expanded_domain}, where we see that the ensemble trained on an expanded domain sees a substantial reduction in RMSE and $L^{\infty}$ norm error.
\begin{table}[h!]
    \centering
      \renewcommand{\arraystretch}{1.5}
    \begin{tabular}{cccc}
        \hline
        \textbf{Function} & \textbf{Training Domain} & \makecell{\textbf{Final RMSE} \\ \bf{on [1.1,4.9]}} & \makecell{\textbf{Final $L^{\infty}$ Norm Error} \\ \bf{on [1.1,4.9]}} \\
        \hline
        \multirow{2}{*}{\textbf{I.13.12}} & [1,5] & 6.061e-08  & 2.569e-06 \\
        & [1.1,4.9] & 3.523e-07  & 1.121e-04 \\
        \hline
    \end{tabular}
    \caption{Validation RMSE and $L^{\infty}$ norm error on $[1.1, 4.9]$ for networks trained on the same domain versus a modestly expanded domain $[1, 5]$. Slight expansion reduces both error metrics.}
    \label{tab:expanded_boundary_table}
\end{table}

To explore the limits of this approach, we conduct a second experiment with a more pronounced domain expansion. Now, the 1,000,000 validation data samples are all generated to have input variables in $[2,4]$. Again, two residual ensembles are trained with 1,000,000 samples: one on $[2, 4]$ and the other on the same expanded domain $[1, 5]$. The final RMSE and $L^{\infty}$ norm error for each ensemble is given in Table \ref{tab:expanded_boundary_table_2}. We now observe that the expanded domain ensemble performs noticeably worse in RMSE than the original domain ensemble. This suggests that expanding the domain dilutes training density per unit volume in the target region. For a fixed training set size, the expanded model must distribute capacity over a larger input space. Therefore, when the expanded region is too large relative to the target domain, the model may underfit the region of interest unless given a proportionally larger training set. 
\begin{table}[h!]
    \centering
      \renewcommand{\arraystretch}{1.5}
    \begin{tabular}{cccc}
        \hline
        \textbf{Function} & \textbf{Training Domain} & \makecell{\textbf{Final RMSE} \\ \bf{on [2,4]}} & \makecell{\textbf{Final $L^{\infty}$ Norm Error} \\ \bf{on [2,4]}} \\
        \hline
        \multirow{2}{*}{\textbf{I.13.12}}  & [1,5] & 2.728e-08  & 1.279e-07 \\
        & [2,4] & 1.883e-10  & 1.057e-07 \\
        \hline
    \end{tabular}
    \caption{Validation RMSE and $L^{\infty}$ norm error on $[2, 4]$ comparing training on the exact target domain versus a much larger domain $[1, 5]$. Overly broad expansion degrades both error metrics due to diluted training density.}
    \label{tab:expanded_boundary_table_2}
\end{table}
\begin{figure}
    \centering
    \includegraphics[width=0.72\linewidth]{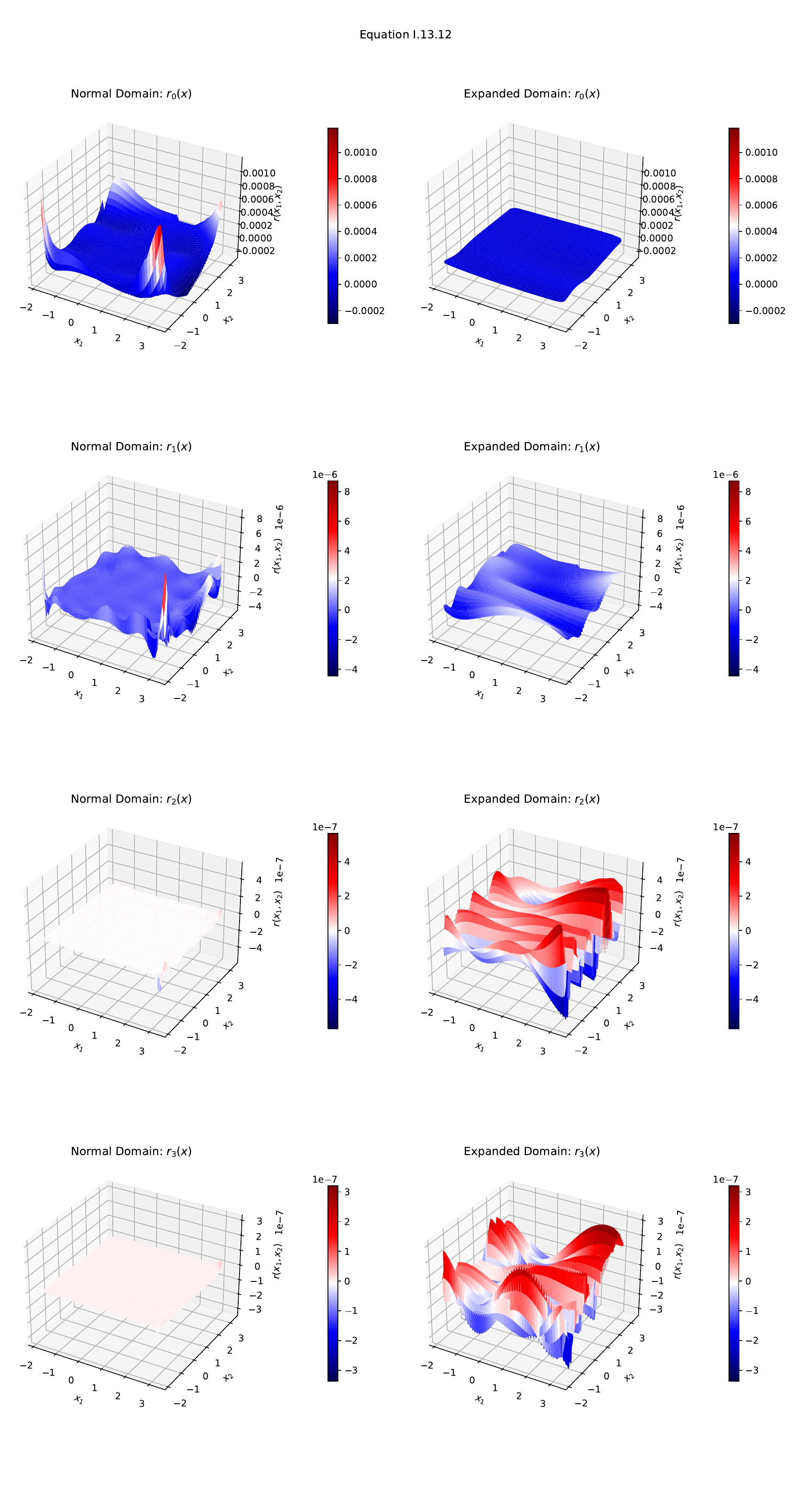}
    \caption{Residuals at each HiPreNet stage for Function I.13.12, evaluated on validation data in $[2, 4]$. The standard network is trained on the same domain, while the expanded network is trained on the much larger domain $[1, 5]$. The expanded domain here performs worse, as the large gap between training and validation domains dilutes training density and degrades accuracy in the target region.}
    \label{fig:expanded_domain_large}
\end{figure}

In summary, expanding the training domain can significantly reduce boundary errors when the expanded region is modest and representative. However, overly broad expansion may lead to underfitting in the target domain unless training density is increased proportionally. Therefore, this technique is most naturally applicable in settings where data generation or collection is cheap and where there is a well-defined boundary of the domain of interest so that a modest expansion can be correctly designed.

\section{High Dimensional ODE Example}\label{section:highdim}
We next evaluate the HiPreNet framework on a higher dimensional problem: a 10-generator 39-bus power system which is a reduced-order model of the New England power grid \cite{athay1979}. The dynamics are described by the 20-dimensional dynamical system
\begin{align*}
    \frac{d\omega_i}{dt} &= \frac{\omega_0}{2H_i}\Bigl(P_m - D\frac{\omega_i-\omega_0}{\omega_0} - E_i^2 G_{ii} \nonumber \\
    &\quad - \sum_{j=1,j\neq i}^{10} E_i E_j \left[B_{ij}\sin(\delta_i - \delta_j) + G_{ij}\cos(\delta_i - \delta_j)\right]\Bigr) \nonumber \\
    \frac{d\delta_i}{dt} &= \omega_i - \omega_0
\end{align*}
for $i=1,\dots,10$, where $\omega_i$ is the rotor speed in rad/s, and $\delta_i$ is the rotor angle in rad. The other parameters are the inertial constant of the generator $H_i$, the synchronous angular frequency $\omega_0 = 2\pi f_0$ in rad/s for an ac power system with frequency $f_0$, the damping coefficient $D$, the mechanical power input from the turbine $P_m$, and the electromotive force or internal voltage of the generator $E_i$. In addition, the mutual admittance $G_{ij} + jB_{ij}$ between $E_i$ and $E_j$ is the $i$th row and $j$th column element of the admittance matrix among all electromotive forces, and $G_{ii}$ is the conductance representing the local load seen from $E_i$. 

We aim to learn a flow map that predicts the state $x(t+5)$ directly from $x(t)$. Neural networks are well suited for this task, as they can approximate high dimensional nonlinear state-to-state mappings and capture the effects of nonlinear interactions over finite time horizons. The five-second prediction horizon makes this task challenging, as the learned flow map must encode long-horizon nonlinear dynamics. High accuracy is therefore essential, since even modest approximation errors may be unacceptable when the flow map is highly nonlinear.

There are several motivating reasons for this choice of a coarse prediction horizon. First, many power system analysis tasks require the system state after several seconds rather than at every smaller interval. Such a task might be assessing whether the system has returned to a near-steady operating point following a disturbance. A 5 second neural network surrogate directly outputs this quantity in a single forward pass, without the storage or processing of intermediate states that a finer resolution surrogate would require, which enables fast inference. Additionally, a finer resolution surrogate applied autoregressively would require many sequential network evaluations, each introducing error that compounds over time. A coarser surrogate avoids this accumulation entirely and achieves a large computational speedup precisely because it bypasses many of these intermediate timesteps in a single evaluation.

To generate data, we numerically solve this system using the Forward Euler method for 500 different initial conditions drawn from the uniform distribution $R$ where
\begin{align*}
    R = &\{x \in \mathcal{R}^{20}; \ -0.4\pi < \delta_i - (\delta_0)_i < 0.4\pi  \\
    &|\omega_i - 120\pi| < 1.5 \ \text{for} \ 1 \leq i \leq 10\}
\end{align*}
using a timestep of $dt=\frac{1}{1000}$ for 20 seconds. From these trajectories, we collect each valid pair of states $\left(x(t),x(t+5)\right)$ for time $t\in[0,15]$, resulting in approximately $7.5$ million samples. 

To learn this flow map, we first train a neural network with four hidden layers of 128 neurons each on 80\% of the total samples. We then apply a single refinement stage to test whether residual correction improves finite-horizon prediction in high dimensional dynamical systems. This was accomplished by training a refinement network with four hidden layers of 256 neurons on the dataset $\left(x(t),r(t) \right)$ where $r(t)$ is the base network's residuals standardized using the residuals' mean and standard deviation. Both networks were optimized using Adam with MSE loss for 20 total epochs. Adam was chosen over BFGS for this problem due to its scalability benefits for higher dimensional systems.

The RMSE and $L^{\infty}$ norm error after evaluating the model on the remaining 20\% of samples are shown in Table \ref{tb:highdim_base}. The refinement network successfully reduces the observed RMSE for $\omega$ by approximately $60\%$ and for $\delta$ by approximately $70\%$, and reduces the $L^{\infty}$ norm error for $\omega$ by approximately $35\%$ and for $\delta$ by approximately $58\%$.

\begin{table}[h]
\centering
\begin{tabular}{lcccc}
\hline
& \textbf{Base $\omega$} & \textbf{Final $\omega$} & \textbf{Base $\delta$} & \textbf{Final $\delta$} \\
\hline
\textbf{RMSE (rad/s or rad)} & 4.130e-03 & 1.670e-03 & 5.647e-03 & 1.484e-03 \\
\textbf{$L^{\infty}$ Error (rad/s or rad)} & 2.765e-01 & 1.802e-01 & 1.242e-01 & 5.252e-02 \\
\hline
\end{tabular}
\caption{Validation RMSE and $L^{\infty}$ norm error for rotor speed $\omega$ and rotor angle $\delta$ for the base and refined networks. Refinement reduces $\omega$ RMSE by approximately $60\%$, $\delta$ RMSE by approximately $70\%$, $\omega$ $L^{\infty}$ error by approximately $35\%$, and $\delta$ $L^{\infty}$ error by approximately $58\%$.}
\label{tb:highdim_base}
\end{table}

\begin{figure}
\begin{center}
\includegraphics[width=\linewidth]{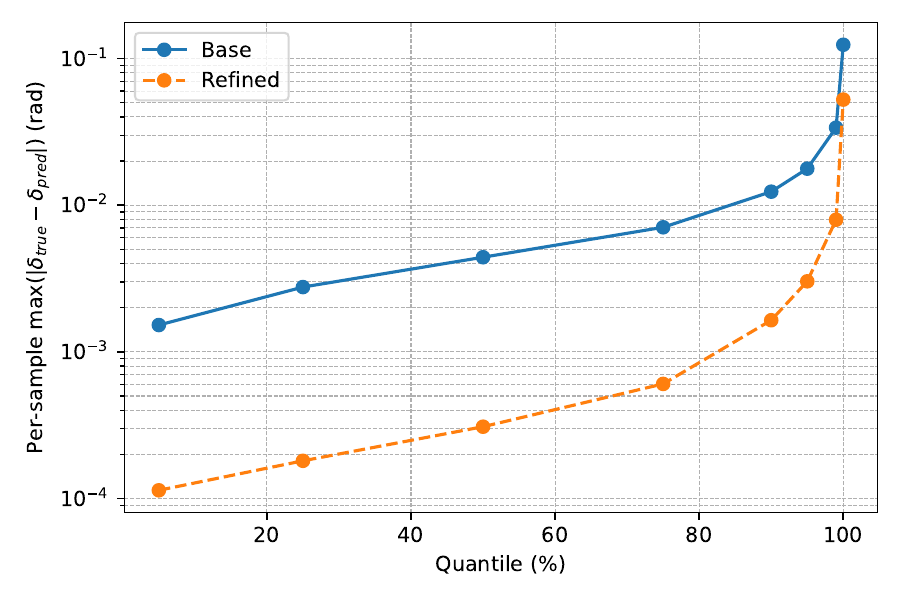}
\caption{Quantiles of the per-sample maximum prediction error across all 10 generators for the rotor angle $\delta$, evaluated on the validation data. Each point represents the error threshold below which that percentage of samples falls. The refined model consistently reduces both typical and worst-case errors relative to the base model.}
\label{fig:quantilesdelta}
\end{center}
\end{figure}

\begin{figure}
\begin{center}
\includegraphics[width=\linewidth]{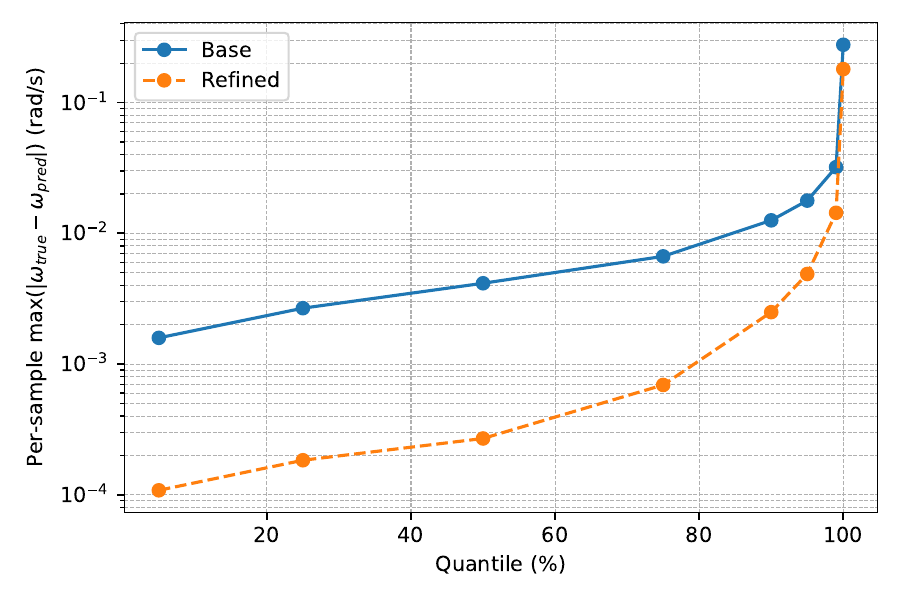}
\caption{Quantiles of the per-sample maximum prediction error across all 10 generators for the rotor speed $\omega$, evaluated on the validation data. Each point represents the error threshold below which that percentage of samples falls. The refined model consistently reduces both typical and worst-case errors relative to the base model.}
\label{fig:quantilesomega}
\end{center}
\end{figure}

Additionally, for each validation sample, with error defined as the $L^{\infty}$ norm error across the 10 generators, we report quantiles of this error for the map from $x(t)$ to $x(t+5)$ in Figure \ref{fig:quantilesdelta} for the rotor angle $\delta$ and in Figure \ref{fig:quantilesomega} for the rotor speed $\omega$. These results indicate that refinement reduces both average error as well as rare, high-magnitude prediction errors for both variables.


\subsection{Targeted Data Augmentation as Weighted Sampling}
From analyzing the residuals of the base model, we observe that the largest ones occur for initial states $x(t)$ where $t \in [0,5]$. This is explained by the dynamics of this system being in a transient state for the first few seconds, before settling into a steady state after approximately 10 seconds. These transient dynamics are more difficult to learn than the steady state dynamics, but the vast majority ($\approx75\%$) of the training dataset is composed of steady state samples. This is similar to the boundary concentration of residuals observed in the low dimensional examples in Section \ref{section:weightedsampling}. However, unlike those cases where the dataset was fixed, here the simulator allows us to generate additional training data cheaply. We therefore apply targeted data generation similar to the approach in Section \ref{targetedsampling}. For this example, we will generate completely new samples concentrated in the high error transient region.

In practice, we generate approximately 2 million more training samples (which are $(x(t),x(t+5))$ pairs), all of which have $t \in [0,5]$. We then augment the original training dataset with these new samples. This shifts the balance of the dataset to be approximately $40\%$ transient samples and $60\%$ steady state samples.

With this augmented training dataset, we compute the residuals using the base model. The base model remains trained solely on the original dataset. These augmented residuals are now used as the targets for the refinement network, which has the same four hidden layers of 256 neurons architecture, and we again optimize it for 20 total epochs using Adam and MSE loss.

The resulting RMSE and $L^{\infty}$ norm error of the augmentation trained network evaluated on the original base validation dataset are given in Table \ref{tb:highdim_aug}. As shown, the $\omega$ RMSE is reduced by approximately $24\%$ and the $\omega$ $L^{\infty}$ norm error by $51\%$ compared to the results from the network trained solely on the base dataset that are reported in Table \ref{tb:highdim_base}. Similarly, the $\delta$ RMSE is reduced by approximately $20\%$ and the $\delta$ $L^{\infty}$ norm error by $23\%$ compared to the results from those tables. Overall, these results suggest that augmenting the dataset with more data concentrated around the large residual samples allows for the refinement network to better correct the base network.


\begin{table}[h]
\centering
\begin{tabular}{lcccc}
\hline
& \textbf{Base Dataset $\omega$} & \textbf{Augmented Dataset $\omega$} & \textbf{Base Dataset $\delta$} & \textbf{Augmented Dataset $\delta$} \\
\hline
\textbf{RMSE} & 1.670e-03 & 1.276e-03 & 1.484e-03 & 1.182e-03 \\
\textbf{$L^{\infty}$ Error} & 1.802e-01 & 8.913e-02 & 5.252e-02 & 4.046e-02 \\
\hline
\end{tabular}
\caption{Validation RMSE and $L^{\infty}$ norm error for rotor speed $\omega$ and rotor angle $\delta$ for the base and refined networks. Refinement reduces $\omega$ RMSE by approximately $24\%$, $\delta$ RMSE by approximately $20\%$, $\omega$ $L^{\infty}$ error by approximately $51\%$, and $\delta$ $L^{\infty}$ error by approximately $23\%$.}
\label{tb:highdim_aug}
\end{table}

\subsection{Flow Map Multiple Iterations Forward}
We next evaluate how well the trained neural networks can approximate the system state at further times. Given the initial state $x(0)$, we autoregressively apply the trained surrogate to predict $x(t+5)$, $x(t+10)$, $x(t+15)$, and 
$x(t+20)$, where each prediction is obtained by feeding the previous prediction back into the model as input.

\begin{figure}
\begin{center}
\includegraphics[width=\linewidth]{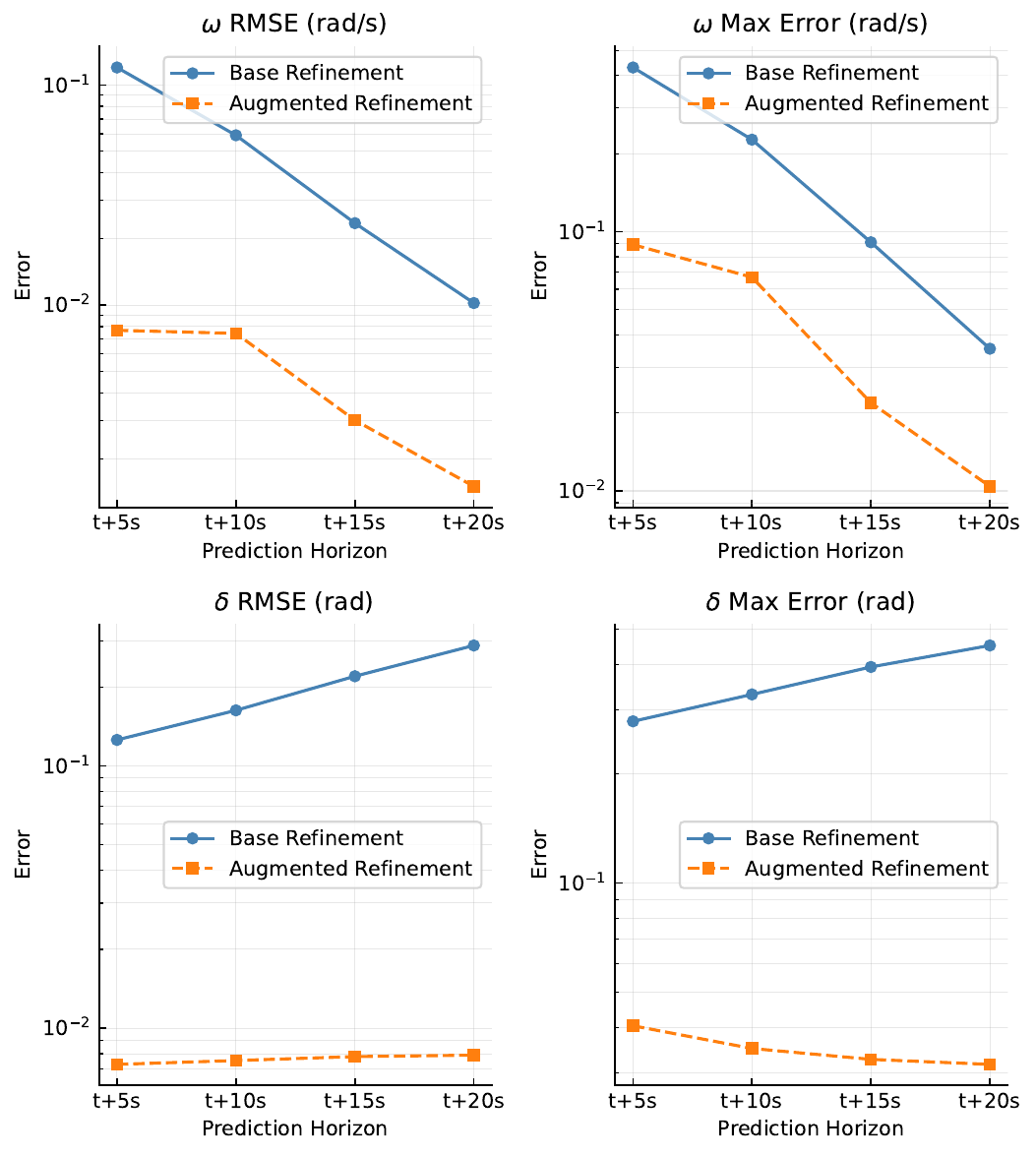}
\caption{RMSE and $L^{\infty}$ norm error for rotor speed $\omega$ and rotor angle $\delta$ as a function of prediction horizon, comparing the base refinement network against the augmented refinement network. For $\omega$, errors decrease with horizon as the system converges to a steady state. For $\delta$, the base refinement network shows growing errors due to error accumulation, while the augmented refinement network maintains stable or decreasing errors across all horizons.}
\label{fig:networkcompare}
\end{center}
\end{figure}

In Figure \ref{fig:networkcompare}, we plot the RMSE and $L^{\infty}$ norm error for both $\omega$ and $\delta$ as functions of the prediction horizon. Results are shown for the base refinement network as well as the augmented refinement network.

For the rotor speed $\omega$, both the RMSE and $L^{\infty}$ norm error decrease as the prediction horizon increases. Physically, this matches with expected behavior, where the power system is stable and rotor speeds converge to a steady state as the transient dynamics decay. There is less variation in $\omega$ to predict as time progresses, and the trained model correctly captures this.

For the rotor angle $\delta$, the results differ between the two refinement networks. Using the base refinement network, we observe that the $\delta$ RMSE and $L^{\infty}$ norm error increase with the prediction horizon. As $\delta$ is an integrated quantity from $\omega$, this behavior is consistent with integrator drift, where small errors in the predicted $\omega$ accumulate into growing $\delta$ errors over successive model steps.

However, when using the augmented refinement network, we see that the $\delta$ errors remain nearly constant or even decrease with the prediction horizon. One likely factor behind this is that the targeted data augmentation substantially reduced early $\omega$ errors, which limited the effects of error accumulation. This can also be explained by the dynamics of the power system, where trajectories converge to a steady state. When the model accurately captures these dynamics, as the augmented refinement network appears to, this attraction toward the true steady state can overcome the error accumulation from $\omega$ errors, resulting in stable or even decreasing $\delta$ error at longer horizons.

\subsection{Neural Network Surrogate Speedup}
By using a neural network surrogate for this model, we also achieve a speedup in wall clock time. We run both the surrogate and Forward Euler solver given 100 different random initial conditions, and compute the mean and standard deviation of the runtime. The surrogate model completes a 5-second prediction in $1.24 \pm 0.20$ ms, compared to $294.87 \pm 2.30$ ms for the Forward Euler solver using a step size of $dt=\frac{1}{1000}$, representing a $238\times$ reduction in computation time. Similarly, the surrogate model completes a 20-second prediction in $11.01 \pm 0.89$ ms, compared to $1033.95 \pm 11.51$ ms for the Forward Euler solver using the same step size, which is a $94\times$ reduction in computation time. This lower speedup reflects the four autoregressive model evaluations required for the 20 second prediction. When using a surrogate we do not obtain any of the intermediate states, so this is only beneficial for applications where only the final state matters.

Additionally, the surrogate naturally supports batch inference for running simulations for multiple initial conditions. This results in a massive speedup versus running simulations sequentially, and this speedup can be further improved by moving inference to run on a GPU. In Table \ref{tb:inferencespeed5s}, we present the mean and per-trajectory runtime across batch sizes of 1, 10, 100 5-second predictions on both a CPU and GPU. The experiments were repeated 50 times. In Table \ref{tb:inferencespeed20s}, we present similar results for 20-second predictions. In both cases, while the GPU is slower for a batch size of 1, it becomes more efficient in per-trajectory time as the batch size grows.

\begin{table}[h]
\centering
\setlength{\tabcolsep}{4pt}
\begin{tabular}{cccc}
\hline
\textbf{Device} & \textbf{Batch Size} & \textbf{Mean $\pm$ Std (ms)} & \textbf{Per-Trajectory (ms)} \\
\hline
\multirow{3}{*}{CPU} & 1   & $1.241 \pm 0.199$ & 1.241 \\
                     & 10  & $2.112 \pm 0.392$ & 0.211 \\
                     & 100 & $13.869 \pm 0.256$ & 0.139 \\
\hline
\multirow{3}{*}{GPU} & 1   & $3.755 \pm 0.392$ & 3.755 \\
                     & 10  & $3.704 \pm 0.281$ & 0.370 \\
                     & 100 & $4.204 \pm 0.129$ & 0.042 \\
\hline
\end{tabular}
\caption{Surrogate model inference time for a single 5-second prediction step across batch sizes on CPU and GPU. Per-trajectory time is the mean divided by batch size.}
\label{tb:inferencespeed5s}
\end{table}

\begin{table}[h]
\centering
\setlength{\tabcolsep}{4pt}
\begin{tabular}{cccc}
\hline
\textbf{Device} & \textbf{Batch Size} & \textbf{Mean $\pm$ Std (ms)} & \textbf{Per-Trajectory (ms)} \\
\hline
\multirow{3}{*}{CPU} & 1   & $11.006 \pm 0.892$ & 11.006 \\
                     & 10  & $21.033 \pm 1.476$ & 2.103  \\
                     & 100 & $61.637 \pm 0.559$ & 0.616  \\
\hline
\multirow{3}{*}{GPU} & 1   & $13.529 \pm 3.909$ & 13.529 \\
                     & 10  & $12.290 \pm 1.167$ & 1.229  \\
                     & 100 & $16.933 \pm 1.069$ & 0.169  \\
\hline
\end{tabular}
\caption{Surrogate model inference time for a 20-second prediction (4 autoregressive steps) across batch sizes on CPU and GPU. Per-trajectory time is the mean divided by batch size.}
\label{tb:inferencespeed20s}
\end{table}

\section{Conclusion}\label{section:conclusion}
Training highly accurate neural networks for scientific and engineering tasks remains challenging due to factors such as non-convex optimization, sensitivity to hyperparameters, and the lack of a structured framework for improving worst-case error. Inspired by the principles of gradient boosting from classical machine learning, our approach decomposes the problem of fitting a complex function into a sequence of simpler tasks, where each stage of training focuses on correcting the residual errors of its predecessor. This method not only reduces the dependence on extensive hyperparameter tuning, but also offers a modular and more scalable path toward model refinement.

Unlike traditional approaches that rely heavily on minimizing average error via mean squared loss, our method also prioritizes control over the $L^{\infty}$ norm error, which is crucial in domains that require safety-critical decision-making. Through extensive experiments on regression problems drawn from the Feynman dataset, we demonstrated that our HiPreNet framework consistently improves both RMSE and $L^{\infty}$ norm error, outperforming standard fully connected networks and even more advanced architectures like Kolmogorov–Arnold Networks (KANs). We additionally showed how the framework can be applied to a higher dimensional power system example, where it allowed for the creation of a fast and accurate neural network surrogate model.


Across both low and high dimensional settings, we observed the consistent theme that the largest errors concentrate in specific, identifiable regions of the input domain. These regions appeared to be near boundaries for the lower dimensional Feynman functions and in the transient regime for the power system model. We addressed this through three related techniques. The first was weighted sampling, which resamples existing data proportionally to residual magnitude. We secondly investigated localized patching as a preliminary technique, which trains a simple neural network to correct a localized spike in the residuals. Thirdly, we tried targeted data augmentation, which generates new samples in the high error region when new data collection is possible. These techniques represent a spectrum of approaches to the same underlying problem, with the appropriate choice depending on whether new data can be generated and whether the high error region has identifiable structure.

In future work, it would be beneficial to explore if there are certain properties of the residuals that can be taken advantage of to determine the optimal size of each refinement network. It would also be worthwhile to modify the framework to handle classification tasks. The HiPreNet framework could also be applied to different architectures such as transformers and neural operators, where investigating whether the residual correction principle remains effective given their different training dynamics appears to be a promising future direction.

\bibliographystyle{abbrv}
\bibliography{Ethan_bib}

\end{document}